\documentclass{article}

\usepackage{microtype}
\usepackage{graphicx}
\usepackage{subfigure}
\usepackage{booktabs} 
\usepackage{array}
\usepackage{hyperref}
\usepackage{amsmath} 
\usepackage{multirow}
\usepackage{makecell}
\usepackage{etoc}
\usepackage{float}  

\usepackage[accepted]{icml2025}

\newcommand{\fieldheading}[1]{\vspace{0.4em}\noindent\textbf{\normalsize #1}}

\usepackage{amsmath}
\usepackage{amssymb}
\usepackage{mathtools}
\usepackage{amsthm}
\definecolor{softorange}{rgb}{1, 0.8, 0.6}

\usepackage[capitalize,noabbrev]{cleveref}

\theoremstyle{plain}

\theoremstyle{definition}

\theoremstyle{remark}

\usepackage[textsize=tiny]{todonotes}

\usepackage[most]{tcolorbox}
\usepackage{xcolor}

\usepackage{etoc}

\usepackage{listings}
\usepackage{courier}
\begin{document}

\twocolumn[
\icmltitle{Agent WARPP: Workflow Adherence via Runtime Parallel Personalization}



\icmlsetsymbol{equal}{*}

\begin{icmlauthorlist}
\icmlauthor{Maria Emilia Mazzolenis}{equal,comp}
\icmlauthor{Ruirui Zhang}{equal,comp}
\end{icmlauthorlist}

\icmlaffiliation{comp}{The Vanguard Group, Charlotte, NC, USA}

\icmlcorrespondingauthor{Maria Emilia Mazzolenis}{emilia\_mazzolenis@vanguard.com}
\icmlcorrespondingauthor{Ruirui Zhang}{ruirui\_zhang@vanguard.com}

\icmlkeywords{Machine Learning, ICML}

\vskip 0.3in
]



\printAffiliationsAndNotice{\icmlEqualContribution} 

\begin{abstract}
Large language models (LLMs) are increasingly applied in task-oriented dialogue (TOD) systems but often struggle with long, conditional workflows that involve external tool calls and depend on user-specific information. We present Workflow Adherence via Runtime Parallel Personalization, or WARPP, a training-free, modular framework that combines multi-agent orchestration with runtime personalization to improve workflow adherence in LLM-based systems. By dynamically pruning conditional branches based on user attributes, the framework reduces reasoning overhead and narrows tool selection at runtime. WARPP deploys a parallelized architecture where a dedicated Personalizer agent operates alongside modular, domain-specific agents to dynamically tailor execution paths in real time. The framework is evaluated across five representative user intents of varying complexity within three domains: banking, flights, and healthcare. Our evaluation leverages synthetic datasets and LLM-powered simulated users to test scenarios with conditional dependencies. Our results demonstrate that WARPP outperforms both the non-personalized method and the ReAct baseline, achieving increasingly larger gains in parameter fidelity and tool accuracy as intent complexity grows, while also reducing average token usage, without any additional training.

\end{abstract}

\section{Introduction}

Large language models (LLMs) have demonstrated remarkable potential across many fields, from customer service \cite{adam2021ai} to finance \cite{zhao2024revolutionizing} and  medicine \cite{thirunavukarasu2023large}. LLMs' ability to comprehend natural language and perform tasks  \cite{brown2020language} has enabled rapid adoption across diverse domains. Despite these advancements, executing long, conditional, and multi-step instructions remains challenging for LLMs, particularly in real-world applications requiring precise procedural reasoning.
Task-oriented dialogue (TOD) systems represent one such domain, where interactions are structured to accomplish specific user goals. These systems typically rely on workflows, which are structured sequences of steps that may involve conditional logic, tool or API calls, and task-specific dependencies. Workflows are designed to fulfill user intents such as booking a flight or updating account details, yet accurately executing these underlying workflows remains challenging for LLMs \cite{Iga2024IntegratingLW}. Moreover, many workflows require real-time personalization based on user-specific factors such as account type, preferences, or interaction history \cite{xu-etal-2024-rethinking}. As the complexity of these workflows increases,  LLMs may experience performance degradation in areas such as reasoning \cite{levy2024same} and retrieval \cite{li2407needlebench}, particularly when dealing with long contexts \cite{liu2024lost}. Recent evaluations have further shown that performance degrades as the overall length of an input increases and that multi-hop reasoning remains a challenge \cite{gavin2024longins}.

A range of strategies have been proposed to address these limitations. Prompting methods such as few-shot learning \cite{brown2020language}, Chain-of-Thought prompting \cite{wei2022chain}, and decomposed prompting \cite{khot2022decomposed} help guide models through complex reasoning by structuring their responses more explicitly. Tool-augmented methods like ReAct \cite{yao2023react} and Toolformer \cite{schick2023toolformer} extend model capabilities by enabling interaction with external APIs during task execution. Beyond prompting, workflow-following can also be enhanced through domain-specific fine-tuning \cite{zhao2024self} or reinforcement learning from human feedback \cite{ouyang2022training}, which help align model behavior with task requirements. Finally, agentic architectures such as CAMEL \cite{li2023camelcommunicativeagentsmind} and AutoGen \cite{wu2023autogen} distribute tasks across specialized agents, narrowing each agent's context window while supporting modular reasoning.

While these strategies improve workflow adherence, they often fall short in ensuring that LLMs can faithfully execute the specific instructions associated with user intents in TOD systems. Studies have shown that LLMs are prone to generating hallucinated outputs and exhibit high sensitivity to prompt formulation \cite{sclar2023quantifying}. Diagnostic analyses further reveal that models continue to struggle with practical tool use: LLMs frequently invoke tools that are not available, apply tools unnecessarily, or fail to detect when a task is unsolvable. They also tend to execute tools in the wrong order, reflecting deeper issues with task understanding and logical sequencing of long instructions \cite{zhang2024toolbehonest}.  Multi-agent systems show promise but also face limitations such as poor conversation management, unclear task specifications, ineffective communication among agents, and premature termination \cite{cemri2025multi}. Additionally, planning errors such as incorrect tool invocation, skipping necessary actions, or failing to respect the logical order of constraints further challenge their ability to execute complex, conditional instructions \cite{ji2024testing}.

Despite extensive work on prompting, tool-augmented methods, and multi-agent systems, LLMs still struggle to follow long, conditional workflows without hallucination or misordering and are limited in their ability to adapt to user-specific contexts without additional training. Consider a healthcare booking system: scheduling a hospital appointment may require retrieving the patient’s profile, screening insurance tier and medical history, validating referrals, checking provider availability with waitlist options, verifying identity with manual-review fallbacks, and assessing urgency for telehealth triage. At each step, user-specific factors such as age, language preference, coverage level, or flagged conditions trigger alternate branches, creating dozens of conditional forks that quickly overwhelm standard LLMs, leading to omitted steps, misordered actions, and unreliable performance.
To address these challenges, we propose \textbf{Agent WARPP}, a training-free, user-adaptive framework that prunes workflows at runtime based on user attributes, and orchestrates parallel agents to execute subtasks efficiently. Our contributions are as follows:

\begin{itemize}
  \item \textbf{Training free personalization.} A framework combining workflow pruning with multi-agent orchestration to improve dialogue adherence.
  \item \textbf{Domain general evaluation.} Synthetic datasets spanning five intents in travel, banking, and healthcare enabling reproducible comparisons.
  \item \textbf{Improved accuracy and efficiency.} Demonstrates improved tool and parameter accuracy while reducing token usage, without additional training.
\end{itemize}

We release our code to support reproducibility and future research at \href{https://github.com/emiliamazzo/WARPP/tree/main}{this} repository.

\section{Related Work}

\subsection{Workflow/Instruction Following in LLMs and Task-Oriented Dialogue}
LLMs have dramatically improved at following natural language instructions, but important challenges remain when instructions are long, complex, or require external actions. Research has explored tool calling, planning, and workflow simplification to address these issues. Tool-augmented methods like Toolformer \cite{schick2023toolformer} expand LLM capabilities by enabling API calls, but introduce new complexities as models must decide which tool to use and when, and such a large action space can lead to hallucinated or misused tool calls \cite{jain2024mitigatingcodellmhallucinations}.
Recent approaches mitigate these errors through prompt engineering, constrained decoding, and explicit fine-tuning \cite{roy-etal-2024-flap, he2024achievingtoolcallingfunctionality, qin2023toolllmfacilitatinglargelanguage}. Reasoning techniques like Chain-of-Thought \cite{wei2022chain}, ReAct \cite{yao2023react}, and Faithful Reasoning \cite{creswell2022faithfulreasoningusinglarge}  improve multi-step task decomposition, but they are often loosely aligned with execution constraints \cite{qiao2024makinglanguagemodelsbetter, qian2024creatortoolcreationdisentangling, qin2024toollearningfoundationmodels}.

In structured domains like customer service or robotics, agents must follow strict workflows with precise sequencing. This has motivated modular, agentic systems \cite{li2025agentorientedplanningmultiagentsystems}, though few adapt workflows dynamically at runtime based on user context.  A growing line of work focuses on workflow simplification; systems like OctoTools \cite{lu2025octotools} and Creator \cite{qian2024creatortoolcreationdisentangling} reduce instruction complexity by trimming or disentangling steps. However, these methods still treat workflow structure as static, and a mechanism to dynamically reshape the instruction path itself remains unexplored.

\subsection{Personalization Methods}
Personalization plays a key role in improving the effectiveness and user satisfaction of intelligent systems, particularly in dialogue, where it enhances task efficiency \cite{zhang-etal-2018-personalizing}, satisfaction \cite{liu2020you}, and engagement \cite{kim2023recent}. Prior work in this area can be broadly grouped into two established categories, model-centric personalization and interaction level personalization, with a third emerging direction focused on automated workflow optimization.

Model-centric approaches personalize what the system says by tailoring language model outputs through in-context learning, memory retrieval, or fine-tuning on user-specific data \cite{mazare2018training, huang2023personalized}. Interaction level personalization adapts what the system does, such as selecting workflows or rerouting tasks, based on static user profiles or predefined rules \cite{bak2019variational}. More recently, these approaches have incorporated real time signals, such as dialogue history, user feedback, or contextual information like location, to support dynamic personalization \cite{liu2020you}. However, even dynamic approaches typically operate within a fixed set of execution paths.

A third emerging direction reframes personalization as a problem of workflow generation and optimization. Rather than relying on predefined workflows, these approaches generate and refine execution strategies through optimization techniques. AFLOW \cite{zhang2025aflowautomatingagenticworkflow} represents workflows as directed graphs, where nodes correspond to language model invocations and edges represent logical control flow, and uses Monte Carlo Tree Search to iteratively improve performance. Although AFLOW demonstrates strong generalization across reasoning tasks, its optimization process occurs offline, resulting in static workflows once discovered. GPTSwarm \cite{zhuge2024languageagentsoptimizablegraphs} aligns with this line of work by using reinforcement learning to optimize both multi-agentic structure and single agent internal decision-making based on performance metrics.

Despite these advances, most systems do not personalize how multi-step tasks are executed during runtime. Execution plans remain fixed once selected or generated, raising questions about how to adapt execution structures dynamically in response to evolving user needs.

\subsection{Agent-Based Architectures}
Multi-agent systems built with LLMs have shown strong potential for tackling complex, multi-step tasks by assigning specialized roles to different agents such as planning, reasoning, or execution \cite{chen2023agentversefacilitatingmultiagentcollaboration, wang2024megaagentpracticalframeworkautonomous}. Architectures like CAMEL and COOPER demonstrate how modular teams of agents can coordinate to decompose problems, collaborate via dialogue, and complete goals more reliably than single-agent baselines \cite{li2023camelcommunicativeagentsmind, cheng2023coopercoordinatingspecializedagents}. These architectures allow each agent’s output to contribute to a shared solution, and some systems even adjust agent behaviors or team composition dynamically during execution \cite{chen2023agentversefacilitatingmultiagentcollaboration}. Other work, such as Sirius \cite{zhao2025siriusselfimprovingmultiagentsystems}, explores how agent cooperation can support self-improvement over time through iterative interaction and reflection.

Parallelization further enhances these architectures by reducing latency and improving task throughput. Executing planned function calls concurrently leads to faster and more accurate outcomes compared to sequential baselines \cite{kim2024llm}. Meanwhile, open-source frameworks such as AutoGen \cite{wu2023autogen} and the OpenAI Agents SDK \cite{openai_agents_sdk} have made it easier to build modular and parallelized agent teams. While multi-agent teams and parallel execution speed up processing, they assume a fixed set of subtasks and few systems fully explore how runtime parallelization and modular agents can work together to support dynamic, personalized workflows.

\section{Methodology}
\subsection{Context and Problem Setup}

We conduct our experiments in a TOD setting, where agents help users complete domain-specific tasks like booking flights or updating account information. Unlike open-domain systems, TOD agents must follow a predefined workflow written in natural language that includes multi-step instructions, conditional logic, tool calls, and decision points. The agent must interpret this workflow, manage the conversation, and execute backend actions. We evaluate system performance based on how accurately it follows these workflows under varying conditions, including branching logic and real-time user inputs.

\subsection{Task Definition}

The task environment includes the following components:

\fieldheading{\textbf{Domain:}} 
We define three domains: Banking, Flights, and Hospital. Each domain has distinct intents, workflows, and tools, and they are designed to demonstrate varying levels of complexity. Banking represents a simple setting with up to five tools and minimal branching; Flights is moderately complex, involving up to ten tools and some conditional logic; and Hospital is the most complex, with more than fifteen tools and deeply branched, context dependent workflows. While these complexity levels illustrate the system’s capabilities, any domain could be framed as simple, intermediate, or complex based on its  workflows and tool requirements.

\fieldheading{\textbf{Intent:}} 
An intent defines a specific task that a user aims to complete within a domain. Banking and Flights domains have two intents each and Hospital has one intent, as shown in Table~\ref{tab:intents}. Agents must recognize the users' intents and resolve them by following the appropriate workflow.

\begin{table}[ht]
\centering
\small
\resizebox{\columnwidth}{!}{%
\begin{tabular}{@{}ll@{}}
\toprule
\textbf{Domain} & \textbf{Intents} \\ \midrule
Banking  & \texttt{updateAddress}, \texttt{withdrawRetirementFunds} \\
Flights  & \texttt{bookFlight}, \texttt{cancelFlight} \\
Hospital & \texttt{processPayment} \\
\bottomrule
\end{tabular}
}
\caption{Domains and their associated intents.}
\label{tab:intents}
\end{table}

\fieldheading{\textbf{Workflow:}} 
A workflow is a sequence of natural language steps and tool calls that define how an agent resolves a given intent. Typically written by subject matter experts, these workflows guide both the conversation and backend actions, often including conditional logic (e.g., “if the address is current, skip verification”). Workflows encode expected task flows and decision points that may trigger different branches based on user context or retrieved data. Example workflows can be found in the Appendix section \ref{sec:workflow-example}.

\fieldheading{\textbf{API/Tools:}} 
Agents interact with a set of domain-specific tools (or APIs) to carry out user intents as defined by the workflow. These tools fall into two functional categories. Information-gathering tools are responsible for retrieving relevant user data from internal systems (e.g., GetCustomerProfile), while execution tools perform the final actions required to complete the task (e.g., SubmitAddressChange). Each tool is associated with a structured \texttt{docstring} that specifies its function description, input parameters, and output format, enabling the agent to reason about tool applicability and correctness during execution. The number of tools per workflow is shown in Table~\ref{tab:task-schema}. To increase realism, all workflows use tools that simulate real APIs by querying customer data, performing computations, accepting user-provided arguments (specified in the data fed to the client LLM), and occasionally injecting latency or failure modes. For complex workflows, we also include direct integrations with real APIs to more closely mirror deployment scenarios.

\begin{table}[ht]
\centering
\resizebox{\columnwidth}{!}{%
\begin{tabular}{@{}lccc@{}}
\toprule
\textbf{Intent} & \textbf{Sample Size} & \textbf{Info Tools} & \textbf{Exec Tools} \\ \midrule
Update Address & 50 & 1 &  4 \\
Withdraw Retirement Funds & 50 & 1 & 2 \\ 
\hline
Book Flight & 50 & 3 & 6 \\
Cancel Flight & 50 & 2 & 6 \\
\hline
Process Payment & 50 & 4 & 15 \\
\bottomrule
\end{tabular}%
}
\caption{Summary of intents with their possible paths and APIS for different domains.}
\label{tab:task-schema}
\end{table}

\section{WARPP Architecture}

\subsection{Architecture}

\begin{figure*}[t]
    \centering
    \makebox[\textwidth][c]{%
        \includegraphics[width=1\linewidth]{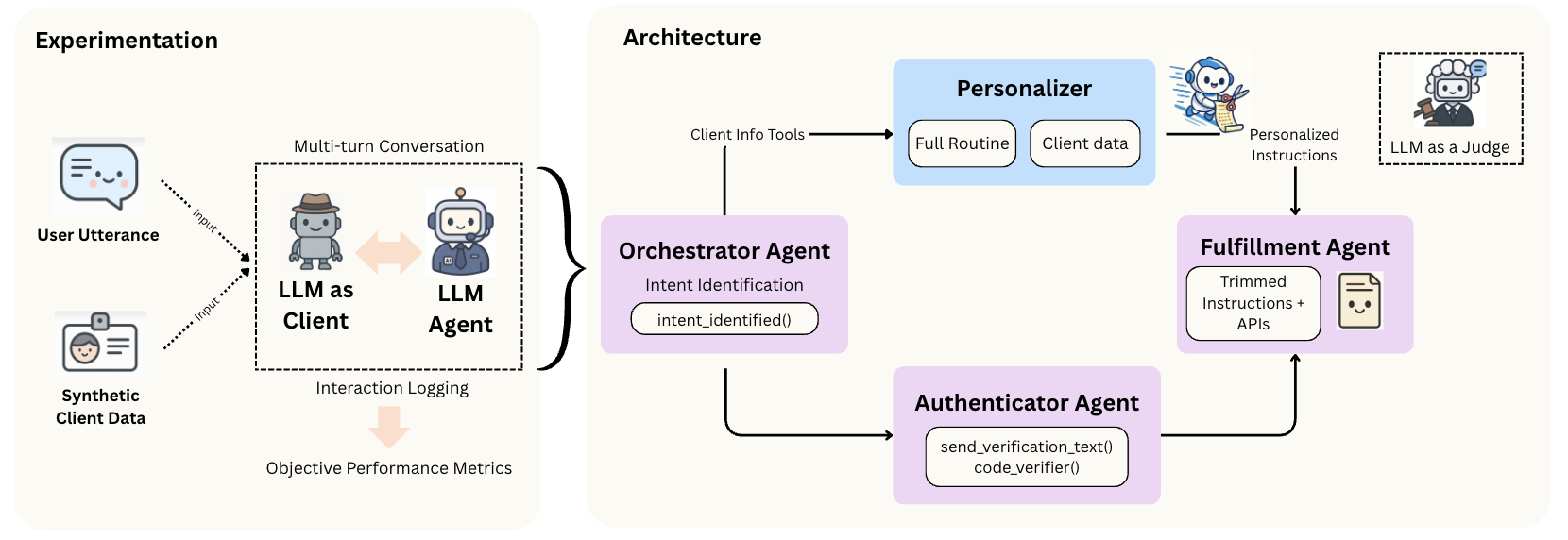}%
    }
    \caption{WARPP Architecture and Experimental Setup}
    \label{fig:enter-label}
\end{figure*}

Our system is built on a multi-agent architecture using OpenAI’s Agents SDK \cite{openai_agents_sdk}, and consists of three core agents: the OrchestratorAgent, the AuthenticatorAgent, and the FulfillmentAgent, along with a parallelized Personalizer agent. These agents coordinate to handle different stages of the dialogue, with a Personalizer agent running in parallel to improve workflow alignment with user context. The architecture can be seen in Figure \ref{fig:enter-label}.

\paragraph{Orchestrator Agent}
The Orchestrator Agent initiates the conversation and identifies the user’s intent. The agent is initialized with domain-specific instructions and a set of supported intents retrieved dynamically from a domain-to-intent mapping. Once the intent is identified, the agent calls a domain-specific tool \verb|intent_identified()|, which returns the full workflow and tool set. Control is passed to the Authenticator Agent, and the Personalizer Agent is launched in parallel. Although the Orchestrator does not participate in fulfillment, its coordination role is critical for modularizing early-stage interactions which are often error-prone in single-agent systems.

\paragraph{Authenticator Agent}
The Authenticator Agent reflects a common requirement in real-world systems for separate identity verification, such as two-factor authentication, while serving as an auxiliary task that runs in parallel with personalization. The Authenticator Agent verifies user identity before any sensitive actions. It uses \verb|send_verification_text()| and \verb|code_verifier()| tools to simulate two-factor authentication with \verb|time.sleep()| calls inserted to approximate realistic response times and better reflect operational latency in production environments. Handoff to the Fulfillment Agent is enforced only after successful verification, ensuring consistency with real-world customer service flows and separation between identity validation and task execution.

\paragraph{Fulfillment Agent}
A single Fulfillment Agent is dynamically configured per intent, avoiding manual duplication. The agent receives personalized or full workflows and tools depending on configuration. In the personalized setup, the Fulfillment Agent is given only the trimmed workflow and the filtered set of execution tools produced by the Personalizer Agent. In the non-personalized setup, the agent receives the full workflow and the complete tool set, which often includes irrelevant branches or unused actions. 

\paragraph{Personalizer agent (Parallelized)}
At the end of the Orchestrator Agent stage, the system sequentially executes all information-gathering tools, which complete almost instantaneously and populate fields required for downstream tool calls. Once complete, the Personalizer applies a three-stage transformation to the full routine using client data from the earlier info-gathering phase, along with a filtered list of available tool calls that excludes the already executed ones:

\begin{itemize}
    \item \textbf{Static Pruning:} Remove branches and tool calls that are incompatible with client attributes, and inline values that can be resolved from client data.
    \item \textbf{Fidelity Preservation:} Retain all outcome branches, including success, failure, and user yes or no responses, around each preserved tool call.
    \item \textbf{Cleanup and Formatting:} Merge descriptive steps and renumber instructions.
\end{itemize}

In addition to the trimmed routine, the Personalizer returns a filtered list of tools required to execute it, containing only those retained after pruning. This output must be finalized before fulfillment begins to ensure consistency between the personalized logic and the tools it depends on. In cases of elevated load or delayed processing, any remaining personalization steps are completed during the brief transition to fulfillment, without compromising logical correctness or user experience.

\paragraph{Workflow Algorithm} Algorithm~\ref{alg:warp} illustrates the end-to-end runtime workflow.

\begin{algorithm}[!ht]
\refstepcounter{algorithm}
\colorbox{softorange}{%
  \parbox{\dimexpr\linewidth-2\fboxsep}{\textbf{Algorithm \thealgorithm} WARPP Execution Protocol}%
}
\label{alg:warp}
\textbf{Inputs:} 

\textit{User attributes} $A = \{a_1, a_2, \dots, a_n\}$ \\
\textit{Full workflow} $W = \{w_1, w_2, \dots, w_k\}$ (each $w_i$ may depend on $A$) \\
\textit{Full tool set} $D = \{d_1, d_2, \dots, d_m\}$ \\
\( D_{\text{info}} \subseteq D, \quad D_{\text{info}} = \{ d_i \in D \mid d_i \text{ is info-gathering tool} \} \)\\
\textit{User utterance} $S$ \\
\textit{Agents}: \textit{Orchestrator} $O_A$, \textit{Authenticator} $A_A$, \textit{Personalizer} $P_A$, \textit{Fulfillment} $F_A$

\textbf{Output:} 

Trimmed workflow $W^*_u$ \& tools $D^*_u$ personalized to user $u$

\textbf{Stage 1: Orchestration and Intent Detection}

\begin{algorithmic}[1]
  \STATE \quad $O_A$ identifies user intent $I$ from $S$
  \STATE \quad Launch $P_A$ asynchronously with $A_A$
\end{algorithmic}

\vspace{0.5em}
\textbf{Stage 2\textnormal{a}: Authenticator (in parallel)}
\begin{algorithmic}[1]
  \STATE \quad Simulated MFA tools called by $A_A$ 
\end{algorithmic}  
\textbf{Stage 2\textnormal{b}: Runtime Personalization (in parallel)}

\begin{algorithmic}[1]
  \STATE \quad \( D_{\text{info}} \) retrieve user attributes \( A \) for user \( u \)
  \STATE \quad $(W^*_u, D^*_u) \gets \textsc{Trim}(W, A)$
\end{algorithmic}

\vspace{0.5em}
\textbf{Stage 3: Fulfillment}

\begin{algorithmic}[1]
  \STATE \quad Pass $(W^*_u, D^*_u)$ to $F_A$
  \STATE \quad $F_A$ follows $W^*_u$ \& executes tools using $D^*_u$
\end{algorithmic}

\end{algorithm}

\subsection{Motivation: Reducing Reasoning Complexity}

To illustrate how the personalization step reduces complexity, consider a full workflow \(W\) that is represented by \(T\) tokens. Each decision point in \(W\) occupies an average of \(t\) tokens, and at each such point the workflow can branch into up to \(b\) paths. Under the assumption that decision logic dominates the workflow, the number of decision points is approximately 
\[
n \approx \frac{T}{t},
\]
and in the worst case,  if all branches are independent, the untrimmed workflow can have up to 
\[
b^n \approx b^{\frac{T}{t}}
\]
distinct execution paths (i.e., complete sequences of tool calls and decisions based on possible user contexts). However, instead of exploring such a large combinatorial space of all potential step-by-step reasoning paths, the WARPP architecture pre-selects the reasonable path. Recent studies have demonstrated that such search space reduction not only streamlines the decision process but also significantly enhances the reasoning accuracy of large language models \cite{yang2025reasonfluxhierarchicalllmreasoning}. By evaluating user attributes early, the Personalizer agent prunes any branch whose condition is unsatisfied, therefore reducing the effective branching factor and shrinking the overall search space. As shown in Algorithm~\ref{alg:warp}, the pruning process requires a single pass over the workflow, yielding a time complexity of \(O(T)\) in terms of token count (or \(O(n)\) in terms of decision points).

Subsequently, we filter out any tools not referenced in the pruned workflow, reducing the tool set from \(|D| = m\) to \(|D^*| \leq m\); this filtering process runs in \(O(m)\) time. Hence, the total complexity of the personalization step is \(O(T + m)\), making it efficient in practice. Moreover, because pruning runs in parallel with user authentication, the personalization does not add significant latency to the overall conversation pipeline.

\section{Experimental Setup}

\subsection{Architectures Evaluated}
We evaluate three distinct system architectures to assess the impact of orchestration and personalization components:

\begin{itemize}
    \item \textbf{ReAct baseline}: A standard ReAct-based pipeline without orchestration \cite{yao2023react}, where a single agent handles all tasks. Internal thoughts are hidden from the client, and observations are integrated into the agent’s context. ReAct offers a widely used single-agent baseline for comparing WARPP’s agentic and personalized approach.

    \item \textbf{WARPP without parallel personalization}: A simplified pipeline following the sequence \texttt{orchestrator} $\rightarrow$ \texttt{authenticator} $\rightarrow$ \texttt{fulfillment}, without parallel execution or personalization modules.
    \item \textbf{Full WARPP with personalizer agent}: The complete WARPP architecture where the \texttt{orchestrator} triggers both the \texttt{personalizer} and \texttt{authenticator} in parallel, followed by \texttt{fulfillment}, enabling personalized and efficient request handling.
\end{itemize}

We ran all three architectures using GPT-4o \cite{hurst2024gpt}, Llama 3 \cite{grattafiori2024llama}, and Claude Sonnet 3.5 \cite{anthropic2024claude35} to evaluate WARPP's performance across models with varying capabilities, cost profiles, and openness (open-source vs. closed-source).

\subsection{Synthetic Data Generation}
Existing TOD datasets like MultiWOZ \citep{budzianowski2020multiwozlargescalemultidomain}, SGD \citep{rastogi2020towards}, STAR \citep{mosig2020star}, ABCD \citep{chen2021abcd}, and SMCalFlow \citep{SMDataflow2020} focus on slot-filling, linear API calls, or narrowly defined workflows. Such datasets lack the complexity needed to study dynamic, multi-step workflows with branching logic and API interdependencies. To address this, we generate synthetic data with rich conditional flows and user-specific variations using a three-step process: (1) define domain-intent-tool mappings, (2) deterministic or LLM-based data generation and (3) manual review.

\subsubsection{Workflow Generation and APIs}
We define JSON schemas for each domain (Banking, Flights, Hospital), specifying intents and their API tools. These guide an LLM in generating initial workflow drafts with conditional logic. To match target complexity, we extensively revise the outputs to refine decision paths and ensure consistent logic. Final workflows are manually validated to reflect realistic, user-dependent interactions.

\subsubsection{Customer Data}
\label{sec:customer-data}
To simulate realistic interactions, we generate synthetic customer profiles using the \texttt{traxgen} \cite{traxgen2025} package. For each intent, we specify the set of user attributes required to complete the associated workflow, along with value distributions for each field. We generate 50 user profiles per intent, ensuring sufficient diversity across scenarios while preserving task relevance. We also generate the initial user utterance for each profile using GPT-4o \cite{hurst2024gpt}.

\subsubsection{Trajectory Ground Truth}
Since workflow trajectories depend on tool call results which may vary across executions, we generate a separate ground-truth trajectory for each experimental run (i.e., per model and architecture). We use the \texttt{traxgen} \cite{traxgen2025} package to construct complete trajectories, including agent actions, tool invocations, and parameter values.

\subsection{Evaluation Metrics}
We evaluate workflow adherence using metrics that capture structural alignment, tool usage accuracy, parameter fidelity, and instruction quality. These metrics are computed at the individual user level and also aggregated across users, intents, and experimental conditions.

\paragraph{Trajectory Accuracy}~
\\
\textit{Exact Match} Whether the predicted sequence of agent and tool calls exactly matches the reference. \\
\textit{Agent Match (Ordered / Any Order):} The percentage of ground-truth agent transitions recovered in the predicted sequence, measured both in order and unordered. \\
\textit{LCS Tools:} The ratio of ground-truth tool calls captured in the longest common subsequence (LCS) of tool names.

\paragraph{Tool Usage}~
\\
\textit{Tool Precision / Recall / F1:} Precision is the fraction of predicted tool calls that are correct; Recall is the fraction of ground-truth tool calls that are recovered; F1 is their harmonic mean. All of such metrics are calculated for tool call excluding parameters. Metrics are reported for the overall system and for fulfillment agents only. \\
\textit{Parameter Match Percentage:} The proportion of correctly filled key-value pairs in tool parameters, matched by tool name.

\paragraph{Interaction Quality}~
\\
\textit{Latency:} Average user-perceived response time, reported overall and for fulfillment agents.

\paragraph{Instruction Quality}~
\\
\textit{Trimmed Workflow Instruction Quality:} Two separate scores from 1 to 5 evaluating the relevance and completeness of each personalized workflow, assigned by an LLM-based judge following a structured rubric (see Section~\ref{sec:llm-judge}).

\subsection{LLM as client}
To scale our experiments, we employ an LLM to simulate the client’s role. The client LLM receives a dynamically constructed prompt (see Appendix Section~\ref{sec:client-prompt}) based on user data, which includes: (1) the target intent (e.g., update address), (2) the utterance to initiate the interaction, and (3) the specific information the client must supply to complete the request (e.g., a new address). By including this information, we ensure the simulated client behaves consistently and remains aligned with the expected task trajectory. We use GPT-4o \cite{hurst2024gpt} as the client model.

\subsection{LLM as judge}
\label{sec:llm-judge}
To assess the quality of trimmed workflows we employ an LLM as a judge \cite{zheng2023judging}. Each prompt provides the judge with: (1) the original full workflow, (2) conditionally relevant attributes from user-specific data (the output from information gathering tools and customer ID), and (3) the personalized workflow generated for that user. The judge assigns two $1$–$5$ scores: \textit{Relevance}, evaluating whether the trimmed routine removes all branches and tools irrelevant to the user given the input data, and \textit{Completeness}, assessing whether it preserves all steps and tool calls needed to fulfill the user’s request. The judge is also required to return a natural language explanation justifying the evaluation. To avoid self-preference biases common in shared generator-judge setups \cite{panickssery2024llm}, we use \texttt{gemini-2.0-flash} for judging, which provides a favorable tradeoff between evaluation quality and cost for rubric-based assessments. To ensure the quality of the judge LLM, 10\% of the  results were randomly sampled  and were independently reviewed by human annotators.

\section{Results}

\subsection{Architecture Effects Across Intent Complexity}
As shown in Table \ref{tab:intent-experiment}, performance generally improves across all five intents when moving from the React baseline to the non-personalized method, and further with the full WARPP configuration. For the simplest intents, the gains are modest: while the non-personalized and full WARPP setups show slight improvements in tool sequencing and parameter accuracy, overall performance remains relatively similar across all three strategies.
For mid complexity intents, both the non-personalized method and WARPP yield more noticeable gains over React, with WARPP often achieving the highest scores. On the most complex intent, the differences between configurations are most pronounced. The non-personalized method recovers much of the degradation observed under React, while full WARPP achieves the strongest results across all metrics. Overall, results follow a consistent pattern: React underperforms relative to both alternatives, with improvements becoming increasingly substantial as task complexity increases.

\begin{table}[t]
\centering
\scriptsize
\begin{tabular}{>{\raggedright\arraybackslash}p{1.1cm}p{0.9cm}*{5}{p{0.7cm}}}
\toprule
Intent & Strategy & \makecell{Exact\\Match} & \makecell{LCS\\Tools} & \makecell{Tool\\F1} & \makecell{Fulfill\\Tool\\F1} & \makecell{Param\\Match\\(\%)} \\
\midrule
\multirow{3}{*}{\makecell[l]{Update\\Address}} & ReAct & 0.73 & 95.98 & 97.43 & 98.54 & 98.32 \\
 & No Per. & 0.89 & \textbf{99.33} & 98.59 & \textbf{98.62} & \textbf{99.12} \\
 & WARPP & \textbf{0.97} & 98.56 & \textbf{99.00} & 97.33 & 98.04 \\
\midrule
\multirow{3}{*}{\makecell[l]{Withdraw\\Retirement\\Funds}} & ReAct & 0.85 & 96.03 & 97.17 & \textbf{98.73} & 96.13 \\
 & No Per. & 0.88 & 97.13 & 96.73 & 94.00 & 97.55 \\
 & WARPP & \textbf{0.91} & \textbf{97.97} & \textbf{98.31} & 93.98 & \textbf{98.14} \\
\midrule
\multirow{3}{*}{\makecell[l]{Book\\Flight}} & ReAct & 0.63 & 96.51 & 96.30 & 95.80 & 97.40 \\
 & No Per. & 0.89 & \textbf{99.35} & 99.11 & 98.57 & \textbf{99.38} \\
 & WARPP & \textbf{0.96} & 99.19 & \textbf{99.47} & \textbf{98.67} & 99.10 \\
\midrule
\multirow{3}{*}{\makecell[l]{Cancel\\Flight}} & ReAct & 0.59 & 93.45 & 92.57 & 90.88 & 92.65 \\
 & No Per. & \textbf{0.94} & \textbf{99.87} & \textbf{99.30} & \textbf{99.09} & \textbf{99.58} \\
 & WARPP & 0.86 & 95.62 & 96.20 & 91.87 & 95.23 \\
\midrule
\multirow{3}{*}{\makecell[l]{Process\\Payment}} & ReAct & 0.16 & 82.93 & 87.95 & 85.89 & 76.19 \\
 & No Per. & 0.16 & 93.04 & 93.52 & 92.40 & 86.66 \\
 & WARPP & \textbf{0.56} & \textbf{94.07} & \textbf{95.46} & \textbf{93.27} & \textbf{92.04} \\
\bottomrule
\end{tabular}
\caption{Performance of strategies (ReAct, No Personalization, WARPP) across five intents. Mean values shown; bold indicates best performance per model-intent pair.}
\label{tab:intent-experiment}
\end{table}

\begin{table}[h]
\centering
\small
\begin{tabular}{lcccc}
\toprule
\textbf{Model} & \textbf{Rel. Avg} & \textbf{Rel. Std} & \textbf{Comp. Avg} & \textbf{Comp. Std} \\
\midrule
GPT-4o        & 4.55 & 0.75 & 4.59 & 0.77 \\
Sonnet        & 4.54 & 0.71 & 4.52 & 0.76 \\
LLaMA-3       & 4.49 & 0.90 & 4.52 & 0.94 \\
\bottomrule
\end{tabular}
\caption{Average relevance and completeness scores (1–5 scale) from LLM judge, with standard deviations.}
\label{tab:judge-scores}
\end{table}

\subsection{Interaction Between Architecture, Intent, \& Model}
As shown in Table \ref{tab:fulltable}, the interaction between model capacity and execution strategy reveals clear patterns across levels of intent complexity. For the simpler intents such as update address, withdraw retirement funds, and book flight, Sonnet consistently performs at or near ceiling across all execution strategies, leaving little room for improvement. In contrast, Llama and, to a lesser extent, GPT start with lower parameter fidelity and tool accuracy under the React baseline but exhibit substantial gains when execution is restructured. Both models improve noticeably under the non-personalized method, with further gains under full WARPP.

The pattern generally holds for cancel flight as well, although Llama shows weaker performance under personalization execution compared to its own non-personalized variant. Manual inspection reveals that Llama sometimes fails to initiate tool use and instead describes intended actions. Despite this, the personalization strategy still outperforms React on most metrics.

On the most complex intent, process payment, all models including Sonnet benefit from personalization. Sonnet shows meaningful gains in parameter consistency and tool coordination, while Llama and GPT achieve their largest improvements when moving from the React baseline to more structured execution. These results suggest that while models with lower baseline performance benefit most from orchestration on simpler tasks, all models, including the strongest, delivered measurable improvements with WARPP.

\subsection{Usage Efficiency}
As shown in Table \ref{tab:fulltable}, average token consumption decreases steadily from the React baseline to the non-parallelized method and further under the full WARPP configuration. Token budgets increase with intent complexity across all models, but the relative savings from execution restructuring remain consistent. GPT uses the fewest tokens, followed by Llama and Sonnet, a ranking that holds for every intent and method. For simple intents, the non-personalized method halves token usage compared to React, with full WARPP providing additional smaller reductions. This pattern continues for intermediate workflows, where full WARPP’s absolute savings grow. For the most complex workflow, process payment, WARPP roughly halves token use versus React and outperforms the non-personalized method by a meaningful margin. Overall, WARPP consistently achieves the greatest token efficiency, especially as dialogue complexity increases.

\subsection{Personalized Workflow Quality}
We evaluated 749 personalized workflows; one was not generated due to a tool-calling error. An LLM judge scored relevance and completeness on a 1–5 scale, with workflows averaging 4.52 (SD = 0.79) and 4.54 (SD = 0.82) respectively, suggesting that the Personalizer agent produces workflows that are generally accurate and well-aligned with user-specific context. As seen in \ref{tab:judge-scores}, GPT-4o achieved the highest mean scores (Rel. = 4.55, Comp. = 4.59), followed closely by Claude Sonnet (Rel. = 4.54, Comp. = 4.52) and LLaMA-3 (Rel. = 4.49, Comp. = 4.52). The lower average and higher variance scores of LLaMA-3 are consistent with its performance in Table~\ref{tab:fulltable}.

\section{Conclusion}
We introduced WARPP, a training free, modular framework that enhances workflow adherence in task oriented dialogue by combining runtime personalization with multi agent orchestration. WARPP dynamically prunes conditional branches based on user attributes through a parallelized Personalizer agent, reducing reasoning complexity while maintaining low latency. Our evaluation demonstrated consistent improvements in parameter fidelity and tool accuracy, particularly as task complexity increases, alongside reductions in token usage compared to baselines.

\begin{table}[H]
\centering
\scriptsize
\begin{tabular}{p{0.8cm}p{0.6cm}p{0.8cm}*{4}{p{0.5cm}}p{0.8cm}}
\toprule
\shortstack{Intent} & \shortstack{Model} & \shortstack{Strategy} & 
\shortstack{Exact\\Match} & \shortstack{LCS\\Tools} & \shortstack{Tool\\F1} & 
\shortstack{Param\\Match} & \shortstack{Token\\Usage} \\
\midrule
\multirow{9}{*}{\makecell{Update\\Address}} & \multirow{3}{*}{Llama} & ReAct & 0.64 & 97.62 & 97.36 & \textbf{99.51} & 4764.6 \\
 &  & No Per. & 0.84 & \textbf{98.00} & 96.97 & 98.00 & 1788.2 \\
 &  & WARPP & \textbf{0.94} & 97.00 & \textbf{98.00} & 95.76 & \textbf{1655.7} \\
\cmidrule(lr){2-8}
 & \multirow{3}{*}{GPT} & ReAct & 0.56 & 90.33 & 94.94 & 95.69 & 3250.7 \\
 &  & No Per. & \textbf{1.00} & \textbf{100.00} & \textbf{100.00} & \textbf{100.00} & 1615.7 \\
 &  & WARPP & 0.98 & 98.67 & 99.00 & 98.35 & \textbf{1096.2} \\
\cmidrule(lr){2-8}
 & \multirow{3}{*}{Sonnet} & ReAct & 0.98 & \textbf{100.00} & \textbf{100.00} & 99.76 & 5017.0 \\
 &  & No Per. & 0.82 & \textbf{100.00} & 98.80 & 99.36 & 2835.7 \\
 &  & WARPP & \textbf{1.00} & \textbf{100.00} & \textbf{100.00} & \textbf{100.00} & \textbf{2104.2} \\
\midrule
\multirow{9}{*}{\makecell{Withdraw\\Retirement\\Funds}} & \multirow{3}{*}{Llama} & ReAct & \textbf{0.94} & \textbf{100.00} & \textbf{99.20} & \textbf{100.00} & 4534.3 \\
 &  & No Per. & 0.88 & \textbf{100.00} & 98.92 & \textbf{100.00} & 1409.9 \\
 &  & WARPP & 0.80 & 95.20 & 96.01 & 95.50 & \textbf{1333.1} \\
\cmidrule(lr){2-8}
 & \multirow{3}{*}{GPT} & ReAct & 0.60 & 88.10 & 92.32 & 88.40 & 3081.4 \\
 &  & No Per. & \textbf{1.00} & \textbf{100.00} & \textbf{100.00} & \textbf{100.00} & 1123.3 \\
 &  & WARPP & 0.96 & 99.50 & 99.71 & 99.67 & \textbf{898.1} \\
\cmidrule(lr){2-8}
 & \multirow{3}{*}{Sonnet} & ReAct & \textbf{1.00} & \textbf{100.00} & \textbf{100.00} & \textbf{100.00} & 4631.1 \\
 &  & No Per. & 0.76 & 91.40 & 91.26 & 92.64 & 2015.7 \\
 &  & WARPP & 0.98 & 99.20 & 99.20 & 99.25 & \textbf{1574.5} \\
\midrule
\multirow{9}{*}{\makecell{Book\\Flight}} & \multirow{3}{*}{Llama} & ReAct & 0.12 & 92.33 & 90.68 & 94.62 & 6561.3 \\
 &  & No Per. & 0.72 & \textbf{98.44} & 97.34 & \textbf{98.14} & 2079.0 \\
 &  & WARPP & \textbf{0.92} & 97.86 & \textbf{98.40} & 97.50 & \textbf{1740.6} \\
\cmidrule(lr){2-8}
 & \multirow{3}{*}{GPT} & ReAct & 0.78 & 97.19 & 98.23 & 97.57 & 4694.1 \\
 &  & No Per. & \textbf{0.96} & 99.60 & \textbf{100.00} & \textbf{100.00} & 1262.7 \\
 &  & WARPP & \textbf{0.96} & \textbf{99.71} & \textbf{100.00} & 99.82 & \textbf{1213.7} \\
\cmidrule(lr){2-8}
 & \multirow{3}{*}{Sonnet} & ReAct & \textbf{1.00} & \textbf{100.00} & \textbf{100.00} & \textbf{100.00} & 7274.1 \\
 &  & No Per. & \textbf{1.00} & \textbf{100.00} & \textbf{100.00} & \textbf{100.00} & 2336.4 \\
 &  & WARPP & \textbf{1.00} & \textbf{100.00} & \textbf{100.00} & \textbf{100.00} & \textbf{2013.8} \\
\midrule
\multirow{9}{*}{\makecell{Cancel\\Flight}} & \multirow{3}{*}{Llama} & ReAct & 0.28 & 88.40 & 86.13 & 88.21 & 6601.5 \\
 &  & No Per. & \textbf{0.88} & \textbf{99.60} & \textbf{98.57} & \textbf{98.75} & 2010.9 \\
 &  & WARPP & 0.58 & 86.85 & 88.60 & 85.69 & \textbf{1746.6} \\
\cmidrule(lr){2-8}
 & \multirow{3}{*}{GPT} & ReAct & 0.66 & 94.22 & 94.99 & 93.44 & 4611.2 \\
 &  & No Per. & 0.98 & \textbf{100.00} & \textbf{100.00} & \textbf{100.00} & 1258.1 \\
 &  & WARPP & \textbf{1.00} & \textbf{100.00} & \textbf{100.00} & \textbf{100.00} & \textbf{1248.7} \\
\cmidrule(lr){2-8}
 & \multirow{3}{*}{Sonnet} & ReAct & 0.82 & 97.73 & 96.60 & 96.31 & 7124.7 \\
 &  & No Per. & 0.96 & \textbf{100.00} & 99.35 & \textbf{100.00} & 2282.4 \\
 &  & WARPP & \textbf{1.00} & \textbf{100.00} & \textbf{100.00} & \textbf{100.00} & \textbf{2108.0} \\
\midrule
\multirow{9}{*}{\makecell{Process\\Payment}} & \multirow{3}{*}{Llama} & ReAct & 0.14 & 63.96 & 72.50 & 56.66 & 6838.6 \\
 &  & No Par. & 0.12 & 90.43 & 89.76 & 84.03 & 3447.2 \\
 &  & WARPP & \textbf{0.36} & \textbf{93.37} & \textbf{94.03} & \textbf{88.80} & \textbf{2645.5} \\
\cmidrule(lr){2-8}
 & \multirow{3}{*}{GPT} & ReAct & 0.16 & 93.89 & 96.53 & 86.69 & 5437.1 \\
 &  & No Par. & 0.18 & \textbf{96.67} & \textbf{97.82} & 89.94 & 2092.0 \\
 &  & WARPP & \textbf{0.56} & 91.71 & 94.18 & \textbf{90.93} & \textbf{1855.1} \\
\cmidrule(lr){2-8}
 & \multirow{3}{*}{Sonnet} & ReAct & 0.18 & 90.93 & 94.84 & 85.21 & 8438.6 \\
 &  & No Par. & 0.18 & 92.01 & 92.99 & 86.00 & 3508.1 \\
 &  & WARPP & \textbf{0.76} & \textbf{97.15} & \textbf{98.16} & \textbf{96.39} & \textbf{2863.2} \\
\bottomrule
\end{tabular}
\caption{Performance across execution strategies (ReAct, No Personalization, WARPP) and LLMs for each intent. Mean values shown; bold indicates best performance per model-intent pair.}
\label{tab:fulltable}
\end{table}

Despite these gains, analysis of trimmed workflows revealed occasional omissions of best practice steps, highlighting opportunities for improving personalization quality. Future work should explore advanced prompt engineering and model allocation strategies such as employing stronger language models for personalization decisions and lighter models for task execution. Additionally, alternative Personalizer designs such as decomposing personalization into multiple calls or using ensemble methods may further enhance trimming fidelity. Overall, WARPP offers a scalable, efficient, and adaptive approach for personalized dialogue systems, and its modular, training free architecture presents a solid foundation for future research and development.

\section{Impact Statement}

This work enables zero-shot, runtime workflow personalization in task-oriented dialogue, improving efficiency. However, tailoring based on user attributes can introduce privacy and fairness risks; we recommend data minimization, user transparency, and ongoing audits to mitigate these concerns.

\section{Limitations}

While our framework shows strong performance, several limitations remain. Most experiments use simulated tool calls, allowing controlled evaluation but not fully capturing real-world factors like latency or failure modes. The Orchestrator Agent is mainly tested on intent identification, with limited disambiguation evaluation due to the simple user queries and intent scope. Our workflows cover diverse branching but would benefit from testing on longer, more nested routines. The benchmark includes five intents across three domains with moderate sample sizes; scalability under high load and adaptability to evolving APIs are left for future work. Additionally, occasional issues with the LLM client simulation sometimes impacted experiment outcomes. Important aspects such as conversation quality, security, privacy, and human-in-the-loop scenarios are outside this study’s scope. These limitations highlight avenues for future research to extend our results.

\newpage
\bibliography{main}

\begin{thebibliography}{58}
\providecommand{\natexlab}[1]{#1}
\providecommand{\url}[1]{\texttt{#1}}
\expandafter\ifx\csname urlstyle\endcsname\relax
  \providecommand{\doi}[1]{doi: #1}\else
  \providecommand{\doi}{doi: \begingroup \urlstyle{rm}\Url}\fi

\bibitem[Adam et~al.(2021)Adam, Wessel, and Benlian]{adam2021ai}
Adam, M., Wessel, M., and Benlian, A.
\newblock Ai-based chatbots in customer service and their effects on user compliance.
\newblock \emph{Electronic Markets}, 31\penalty0 (2):\penalty0 427--445, 2021.

\bibitem[Andreas et~al.(2020)Andreas, Bufe, Burkett, Chen, Clausman, Crawford, Crim, DeLoach, Dorner, Eisner, et~al.]{SMDataflow2020}
Andreas, J., Bufe, J., Burkett, D., Chen, C., Clausman, J., Crawford, J., Crim, K., DeLoach, J., Dorner, L., Eisner, J., et~al.
\newblock Task-oriented dialogue as dataflow synthesis.
\newblock \emph{Transactions of the Association for Computational Linguistics}, 8:\penalty0 556--571, 2020.

\bibitem[Anthropic(2024)]{anthropic2024claude35}
Anthropic.
\newblock Introducing claude 3.5 sonnet.
\newblock \url{https://www.anthropic.com/news/claude-3-5-sonnet}, 2024.
\newblock Accessed: 2025-05-27.

\bibitem[Bak \& Oh(2019)Bak and Oh]{bak2019variational}
Bak, J. and Oh, A.
\newblock Variational hierarchical user-based conversation model.
\newblock In \emph{Proceedings of the 2019 Conference on Empirical Methods in Natural Language Processing and the 9th International Joint Conference on Natural Language Processing (EMNLP-IJCNLP)}, pp.\  1941--1950, 2019.

\bibitem[Brown et~al.(2020)Brown, Mann, Ryder, Subbiah, Kaplan, Dhariwal, Neelakantan, Shyam, Sastry, Askell, et~al.]{brown2020language}
Brown, T., Mann, B., Ryder, N., Subbiah, M., Kaplan, J.~D., Dhariwal, P., Neelakantan, A., Shyam, P., Sastry, G., Askell, A., et~al.
\newblock Language models are few-shot learners.
\newblock \emph{Advances in Neural Information Processing Systems}, 33:\penalty0 1877--1901, 2020.

\bibitem[Budzianowski et~al.(2020)Budzianowski, Wen, Tseng, Casanueva, Ultes, Ramadan, and Gašić]{budzianowski2020multiwozlargescalemultidomain}
Budzianowski, P., Wen, T.-H., Tseng, B.-H., Casanueva, I., Ultes, S., Ramadan, O., and Gašić, M.
\newblock Multiwoz -- a large-scale multi-domain wizard-of-oz dataset for task-oriented dialogue modelling.
\newblock \emph{arXiv preprint arXiv:1810.00278}, 2020.

\bibitem[Cemri et~al.(2025)Cemri, Pan, Yang, Agrawal, Chopra, Tiwari, Keutzer, Parameswaran, Klein, Ramchandran, et~al.]{cemri2025multi}
Cemri, M., Pan, M.~Z., Yang, S., Agrawal, L.~A., Chopra, B., Tiwari, R., Keutzer, K., Parameswaran, A., Klein, D., Ramchandran, K., et~al.
\newblock Why do multi-agent {LLM} systems fail?
\newblock \emph{arXiv preprint arXiv:2503.13657}, 2025.

\bibitem[Chen et~al.(2021)Chen, Chen, Yang, Lin, and Yu]{chen2021abcd}
Chen, D., Chen, H., Yang, Y., Lin, A., and Yu, Z.
\newblock Action-based conversations dataset: A corpus for building more in-depth task-oriented dialogue systems.
\newblock \emph{arXiv preprint arXiv:2104.00783}, 2021.

\bibitem[Chen et~al.(2023)Chen, Su, Zuo, Yang, Yuan, Qian, Chan, Qin, Lu, Xie, et~al.]{chen2023agentversefacilitatingmultiagentcollaboration}
Chen, W., Su, Y., Zuo, J., Yang, C., Yuan, C., Qian, C., Chan, C.-M., Qin, Y., Lu, Y., Xie, R., et~al.
\newblock Agentverse: Facilitating multi-agent collaboration and exploring emergent behaviors in agents.
\newblock \emph{arXiv preprint arXiv:2308.10848}, 2\penalty0 (4):\penalty0 6, 2023.

\bibitem[Chen et~al.(2024)Chen, Nishida, Nakayama, and Matsumoto]{kim2023recent}
Chen, Y.-P., Nishida, N., Nakayama, H., and Matsumoto, Y.
\newblock Recent trends in personalized dialogue generation: A review of datasets, methodologies, and evaluations.
\newblock \emph{arXiv preprint arXiv:2405.17974}, 2024.

\bibitem[Cheng et~al.(2024)Cheng, Liu, Wang, Leong, Ouyang, Li, Wu, and Zheng]{cheng2023coopercoordinatingspecializedagents}
Cheng, Y., Liu, W., Wang, J., Leong, C.~T., Ouyang, Y., Li, W., Wu, X., and Zheng, Y.
\newblock Cooper: Coordinating specialized agents towards a complex dialogue goal.
\newblock In \emph{Proceedings of the AAAI Conference on Artificial Intelligence}, volume~38, pp.\  17853--17861, 2024.

\bibitem[Creswell \& Shanahan(2022)Creswell and Shanahan]{creswell2022faithfulreasoningusinglarge}
Creswell, A. and Shanahan, M.
\newblock Faithful reasoning using large language models.
\newblock \emph{arXiv preprint arXiv:2208.14271}, 2022.

\bibitem[Gavin et~al.(2024)Gavin, Zheng, Liu, Que, Wang, Yang, Zhang, Huang, Chen, and Zhang]{gavin2024longins}
Gavin, S., Zheng, T., Liu, J., Que, Q., Wang, N., Yang, J., Zhang, C., Huang, W., Chen, W., and Zhang, G.
\newblock Longins: A challenging long-context instruction-based exam for {LLMs}.
\newblock \emph{arXiv preprint arXiv:2406.17588}, 2024.

\bibitem[Grattafiori et~al.(2024)Grattafiori, Dubey, Jauhri, Pandey, Kadian, Al-Dahle, Letman, Mathur, Schelten, Vaughan, et~al.]{grattafiori2024llama}
Grattafiori, A., Dubey, A., Jauhri, A., Pandey, A., Kadian, A., Al-Dahle, A., Letman, A., Mathur, A., Schelten, A., Vaughan, A., et~al.
\newblock The llama 3 herd of models.
\newblock \emph{arXiv preprint arXiv:2407.21783}, 2024.

\bibitem[He(2024)]{he2024achievingtoolcallingfunctionality}
He, S.
\newblock Achieving tool calling functionality in {LLMs} using only prompt engineering without fine-tuning.
\newblock \emph{arXiv preprint arXiv:2407.04997}, 2024.

\bibitem[Huang et~al.(2023)Huang, Zhang, Ko, Liu, Wu, Wang, and Tang]{huang2023personalized}
Huang, Q., Zhang, Y., Ko, T., Liu, X., Wu, B., Wang, W., and Tang, H.
\newblock Personalized dialogue generation with persona-adaptive attention.
\newblock In \emph{Proceedings of the AAAI Conference on Artificial Intelligence}, volume~37, pp.\  12916--12923, 2023.

\bibitem[Hurst et~al.(2024)Hurst, Lerer, Goucher, Perelman, Ramesh, Clark, Ostrow, Welihinda, Hayes, Radford, et~al.]{hurst2024gpt}
Hurst, A., Lerer, A., Goucher, A.~P., Perelman, A., Ramesh, A., Clark, A., Ostrow, A., Welihinda, A., Hayes, A., Radford, A., et~al.
\newblock {GPT}-4o system card.
\newblock \emph{arXiv preprint arXiv:2410.21276}, 2024.

\bibitem[Iga(2024)]{Iga2024IntegratingLW}
Iga, V. I.-R.
\newblock Integrating {LLMs} with knowledge graphs-enhanced task-oriented dialogue systems.
\newblock In \emph{International Conference on Advanced Information Systems Engineering}, pp.\  40--51, 2024.

\bibitem[Jain et~al.(2024)Jain, Kwiatkowski, Ray, Ramanathan, and Kumar]{jain2024mitigatingcodellmhallucinations}
Jain, N., Kwiatkowski, R., Ray, B., Ramanathan, M.~K., and Kumar, V.
\newblock On mitigating code {LLM} hallucinations with api documentation.
\newblock \emph{arXiv preprint arXiv:2407.09726}, 2024.

\bibitem[Ji et~al.(2024)Ji, Wu, Ma, Li, and Wang]{ji2024testing}
Ji, Z., Wu, D., Ma, P., Li, Z., and Wang, S.
\newblock Testing and understanding erroneous planning in {LLM} agents through synthesized user inputs.
\newblock \emph{arXiv preprint arXiv:2404.17833}, 2024.

\bibitem[Khot et~al.(2022)Khot, Trivedi, Finlayson, Fu, Richardson, Clark, and Sabharwal]{khot2022decomposed}
Khot, T., Trivedi, H., Finlayson, M., Fu, Y., Richardson, K., Clark, P., and Sabharwal, A.
\newblock Decomposed prompting: A modular approach for solving complex tasks.
\newblock \emph{arXiv preprint arXiv:2210.02406}, 2022.

\bibitem[Kim et~al.(2024)Kim, Moon, Tabrizi, Lee, Mahoney, Keutzer, and Gholami]{kim2024llm}
Kim, S., Moon, S., Tabrizi, R., Lee, N., Mahoney, M.~W., Keutzer, K., and Gholami, A.
\newblock An {LLM} compiler for parallel function calling.
\newblock In \emph{Forty-first International Conference on Machine Learning}, 2024.

\bibitem[Levy et~al.(2024)Levy, Jacoby, and Goldberg]{levy2024same}
Levy, M., Jacoby, A., and Goldberg, Y.
\newblock Same task, more tokens: the impact of input length on the reasoning performance of large language models.
\newblock \emph{arXiv preprint arXiv:2402.14848}, 2024.

\bibitem[Li et~al.(2024{\natexlab{a}})Li, Xie, Li, Tsung, Ding, and Li]{li2025agentorientedplanningmultiagentsystems}
Li, A., Xie, Y., Li, S., Tsung, F., Ding, B., and Li, Y.
\newblock Agent-oriented planning in multi-agent systems.
\newblock \emph{arXiv preprint arXiv:2410.02189}, 2024{\natexlab{a}}.

\bibitem[Li et~al.(2023)Li, Hammoud, Itani, Khizbullin, and Ghanem]{li2023camelcommunicativeagentsmind}
Li, G., Hammoud, H., Itani, H., Khizbullin, D., and Ghanem, B.
\newblock Camel: Communicative agents for" mind" exploration of large language model society.
\newblock \emph{Advances in Neural Information Processing Systems}, 36:\penalty0 51991--52008, 2023.

\bibitem[Li et~al.(2024{\natexlab{b}})Li, Zhang, Liu, and Chen]{li2407needlebench}
Li, M., Zhang, S., Liu, Y., and Chen, K.
\newblock Needlebench: Can {LLMs} do retrieval and reasoning in 1 million context window?, 2024.
\newblock \emph{arXiv e-prints}, pp.\  arXiv--2407, 2024{\natexlab{b}}.

\bibitem[Liu et~al.(2024)Liu, Lin, Hewitt, Paranjape, Bevilacqua, Petroni, and Liang]{liu2024lost}
Liu, N.~F., Lin, K., Hewitt, J., Paranjape, A., Bevilacqua, M., Petroni, F., and Liang, P.
\newblock Lost in the middle: How language models use long contexts.
\newblock \emph{Transactions of the Association for Computational Linguistics}, 12:\penalty0 157--173, 2024.

\bibitem[Liu et~al.(2020)Liu, Chen, Chen, Lou, Chen, Zhou, and Zhang]{liu2020you}
Liu, Q., Chen, Y., Chen, B., Lou, J.-G., Chen, Z., Zhou, B., and Zhang, D.
\newblock You impress me: Dialogue generation via mutual persona perception.
\newblock \emph{arXiv preprint arXiv:2004.05388}, 2020.

\bibitem[Lu et~al.(2025)Lu, Chen, Liu, Thapa, Boen, and Zou]{lu2025octotools}
Lu, P., Chen, B., Liu, S., Thapa, R., Boen, J., and Zou, J.
\newblock Octotools: An agentic framework with extensible tools for complex reasoning.
\newblock \emph{arXiv preprint arXiv:2502.11271}, 2025.

\bibitem[Mazar{\'e} et~al.(2018)Mazar{\'e}, Humeau, Raison, and Bordes]{mazare2018training}
Mazar{\'e}, P.-E., Humeau, S., Raison, M., and Bordes, A.
\newblock Training millions of personalized dialogue agents.
\newblock \emph{arXiv preprint arXiv:1809.01984}, 2018.

\bibitem[Mosig et~al.(2020)Mosig, Mehri, and Kober]{mosig2020star}
Mosig, J.~E., Mehri, S., and Kober, T.
\newblock Star: A schema-guided dialog dataset for transfer learning.
\newblock \emph{arXiv preprint arXiv:2010.11853}, 2020.

\bibitem[OpenAI(2025)]{openai_agents_sdk}
OpenAI.
\newblock Openai agents sdk, 2025.
\newblock URL \url{https://github.com/openai/openai-agents-python}.
\newblock Accessed: April 1, 2025.

\bibitem[Ouyang et~al.(2022)Ouyang, Wu, Jiang, Almeida, Wainwright, Mishkin, Zhang, Agarwal, Slama, Ray, et~al.]{ouyang2022training}
Ouyang, L., Wu, J., Jiang, X., Almeida, D., Wainwright, C., Mishkin, P., Zhang, C., Agarwal, S., Slama, K., Ray, A., et~al.
\newblock Training language models to follow instructions with human feedback.
\newblock \emph{Advances in Neural Information Processing Systems}, 35:\penalty0 27730--27744, 2022.

\bibitem[Panickssery et~al.(2024)Panickssery, Bowman, and Feng]{panickssery2024llm}
Panickssery, A., Bowman, S., and Feng, S.
\newblock {LLM} evaluators recognize and favor their own generations.
\newblock \emph{Advances in Neural Information Processing Systems}, 37:\penalty0 68772--68802, 2024.

\bibitem[Qian et~al.(2023)Qian, Han, Fung, Qin, Liu, and Ji]{qian2024creatortoolcreationdisentangling}
Qian, C., Han, C., Fung, Y.~R., Qin, Y., Liu, Z., and Ji, H.
\newblock {CREATOR}: Tool creation for disentangling abstract and concrete reasoning of large language models.
\newblock \emph{arXiv preprint arXiv:2305.14318}, 2023.

\bibitem[Qiao et~al.(2023)Qiao, Gui, Lv, Jia, Chen, and Zhang]{qiao2024makinglanguagemodelsbetter}
Qiao, S., Gui, H., Lv, C., Jia, Q., Chen, H., and Zhang, N.
\newblock Making language models better tool learners with execution feedback.
\newblock \emph{arXiv preprint arXiv:2305.13068}, 2023.

\bibitem[Qin et~al.(2023)Qin, Liang, Ye, Zhu, Yan, Lu, Lin, Cong, Tang, Qian, et~al.]{qin2023toolllmfacilitatinglargelanguage}
Qin, Y., Liang, S., Ye, Y., Zhu, K., Yan, L., Lu, Y., Lin, Y., Cong, X., Tang, X., Qian, B., et~al.
\newblock Toolllm: Facilitating large language models to master 16000+ real-world apis.
\newblock \emph{arXiv preprint arXiv:2307.16789}, 2023.

\bibitem[Qin et~al.(2024)Qin, Hu, Lin, Chen, Ding, Cui, Zeng, Zhou, Huang, Xiao, et~al.]{qin2024toollearningfoundationmodels}
Qin, Y., Hu, S., Lin, Y., Chen, W., Ding, N., Cui, G., Zeng, Z., Zhou, X., Huang, Y., Xiao, C., et~al.
\newblock Tool learning with foundation models.
\newblock \emph{ACM Computing Surveys}, 57\penalty0 (4):\penalty0 1--40, 2024.

\bibitem[Rastogi et~al.(2020)Rastogi, Zang, Sunkara, Gupta, and Khaitan]{rastogi2020towards}
Rastogi, A., Zang, X., Sunkara, S., Gupta, R., and Khaitan, P.
\newblock Towards scalable multi-domain conversational agents: The schema-guided dialogue dataset.
\newblock In \emph{Proceedings of the AAAI Conference on Artificial Intelligence}, volume~34, pp.\  8689--8696, 2020.

\bibitem[Roy et~al.(2024)Roy, Sengupta, Bonadiman, Mansour, and Gupta]{roy-etal-2024-flap}
Roy, S., Sengupta, S., Bonadiman, D., Mansour, S., and Gupta, A.
\newblock Flap: Flow-adhering planning with constrained decoding in {LLMs}.
\newblock \emph{arXiv preprint arXiv:2403.05766}, 2024.

\bibitem[Schick et~al.(2023)Schick, Dwivedi-Yu, Dess{\`\i}, Raileanu, Lomeli, Hambro, Zettlemoyer, Cancedda, and Scialom]{schick2023toolformer}
Schick, T., Dwivedi-Yu, J., Dess{\`\i}, R., Raileanu, R., Lomeli, M., Hambro, E., Zettlemoyer, L., Cancedda, N., and Scialom, T.
\newblock Toolformer: Language models can teach themselves to use tools.
\newblock \emph{Advances in Neural Information Processing Systems}, 36:\penalty0 68539--68551, 2023.

\bibitem[Sclar et~al.(2023)Sclar, Choi, Tsvetkov, and Suhr]{sclar2023quantifying}
Sclar, M., Choi, Y., Tsvetkov, Y., and Suhr, A.
\newblock Quantifying language models' sensitivity to spurious features in prompt design or: How i learned to start worrying about prompt formatting.
\newblock \emph{arXiv preprint arXiv:2310.11324}, 2023.

\bibitem[Thirunavukarasu et~al.(2023)Thirunavukarasu, Ting, Elangovan, Gutierrez, Tan, and Ting]{thirunavukarasu2023large}
Thirunavukarasu, A.~J., Ting, D. S.~J., Elangovan, K., Gutierrez, L., Tan, T.~F., and Ting, D. S.~W.
\newblock Large language models in medicine.
\newblock \emph{Nature medicine}, 29\penalty0 (8):\penalty0 1930--1940, 2023.

\bibitem[Traxgen()]{traxgen2025}
Traxgen.
\newblock Traxgen: Trajectory ground truth generator for agentic frameworks.
\newblock \url{https://pypi.org/project/traxgen/}, 2025.
\newblock Python package (version 0.1.5).

\bibitem[Wang et~al.(2024)Wang, Wang, Li, Liang, and He]{wang2024megaagentpracticalframeworkautonomous}
Wang, Q., Wang, T., Li, Q., Liang, J., and He, B.
\newblock Megaagent: A practical framework for autonomous cooperation in large-scale {LLM} agent systems.
\newblock \emph{arXiv e-prints}, pp.\  arXiv--2408, 2024.

\bibitem[Wei et~al.(2022)Wei, Wang, Schuurmans, Bosma, Xia, Chi, Le, Zhou, et~al.]{wei2022chain}
Wei, J., Wang, X., Schuurmans, D., Bosma, M., Xia, F., Chi, E., Le, Q.~V., Zhou, D., et~al.
\newblock Chain-of-thought prompting elicits reasoning in large language models.
\newblock \emph{Advances in Neural Information Processing Systems}, 35:\penalty0 24824--24837, 2022.

\bibitem[Wu et~al.(2023)Wu, Bansal, Zhang, Wu, Li, Zhu, Jiang, Zhang, Zhang, Liu, et~al.]{wu2023autogen}
Wu, Q., Bansal, G., Zhang, J., Wu, Y., Li, B., Zhu, E., Jiang, L., Zhang, X., Zhang, S., Liu, J., et~al.
\newblock Autogen: Enabling next-gen {LLM} applications via multi-agent conversation.
\newblock \emph{arXiv preprint arXiv:2308.08155}, 2023.

\bibitem[Xu et~al.(2024)Xu, Mao, Yang, Sun, and Huang]{xu-etal-2024-rethinking}
Xu, H.-D., Mao, X.-L., Yang, P., Sun, F., and Huang, H.-Y.
\newblock Rethinking task-oriented dialogue systems: From complex modularity to zero-shot autonomous agent.
\newblock In \emph{Proceedings of the 62nd Annual Meeting of the Association for Computational Linguistics (Volume 1: Long Papers)}, pp.\  2748--2763, 2024.

\bibitem[Yang et~al.(2025)Yang, Yu, Cui, and Wang]{yang2025reasonfluxhierarchicalllmreasoning}
Yang, L., Yu, Z., Cui, B., and Wang, M.
\newblock Reasonflux: Hierarchical {LLM} reasoning via scaling thought templates.
\newblock \emph{arXiv preprint arXiv:2502.06772}, 2025.

\bibitem[Yao et~al.(2023)Yao, Zhao, Yu, Du, Shafran, Narasimhan, and Cao]{yao2023react}
Yao, S., Zhao, J., Yu, D., Du, N., Shafran, I., Narasimhan, K., and Cao, Y.
\newblock React: Synergizing reasoning and acting in language models.
\newblock In \emph{International Conference on Learning Representations (ICLR)}, 2023.

\bibitem[Zhang et~al.(2024{\natexlab{a}})Zhang, Xiang, Yu, Teng, Chen, Chen, Zhuge, Cheng, Hong, Wang, et~al.]{zhang2025aflowautomatingagenticworkflow}
Zhang, J., Xiang, J., Yu, Z., Teng, F., Chen, X., Chen, J., Zhuge, M., Cheng, X., Hong, S., Wang, J., et~al.
\newblock Aflow: Automating agentic workflow generation.
\newblock \emph{arXiv preprint arXiv:2410.10762}, 2024{\natexlab{a}}.

\bibitem[Zhang et~al.(2018)Zhang, Dinan, Urbanek, Szlam, Kiela, and Weston]{zhang-etal-2018-personalizing}
Zhang, S., Dinan, E., Urbanek, J., Szlam, A., Kiela, D., and Weston, J.
\newblock Personalizing dialogue agents: I have a dog, do you have pets too?
\newblock \emph{arXiv preprint arXiv:1801.07243}, 2018.

\bibitem[Zhang et~al.(2024{\natexlab{b}})Zhang, Chen, Wang, Liu, Yang, Shi, Zhu, Lin, Wan, Yang, et~al.]{zhang2024toolbehonest}
Zhang, Y., Chen, J., Wang, J., Liu, Y., Yang, C., Shi, C., Zhu, X., Lin, Z., Wan, H., Yang, Y., et~al.
\newblock Toolbehonest: A multi-level hallucination diagnostic benchmark for tool-augmented large language models.
\newblock \emph{arXiv preprint arXiv:2406.20015}, 2024{\natexlab{b}}.

\bibitem[Zhao et~al.(2024{\natexlab{a}})Zhao, Jia, Viswanathan, Wu, and Neubig]{zhao2024self}
Zhao, C., Jia, X., Viswanathan, V., Wu, T., and Neubig, G.
\newblock Self-guide: Better task-specific instruction following via self-synthetic finetuning.
\newblock \emph{arXiv preprint arXiv:2407.12874}, 2024{\natexlab{a}}.

\bibitem[Zhao et~al.(2024{\natexlab{b}})Zhao, Liu, Wu, Li, Yang, Shu, Xu, Dai, Zhao, Mai, et~al.]{zhao2024revolutionizing}
Zhao, H., Liu, Z., Wu, Z., Li, Y., Yang, T., Shu, P., Xu, S., Dai, H., Zhao, L., Mai, G., et~al.
\newblock Revolutionizing finance with {LLMs} : An overview of applications and insights.
\newblock \emph{arXiv preprint arXiv:2401.11641}, 2024{\natexlab{b}}.

\bibitem[Zhao et~al.(2025)Zhao, Yuksekgonul, Wu, and Zou]{zhao2025siriusselfimprovingmultiagentsystems}
Zhao, W., Yuksekgonul, M., Wu, S., and Zou, J.
\newblock Sirius: Self-improving multi-agent systems via bootstrapped reasoning.
\newblock \emph{arXiv preprint arXiv:2502.04780}, 2025.

\bibitem[Zheng et~al.(2023)Zheng, Chiang, Sheng, Zhuang, Wu, Zhuang, Lin, Li, Li, Xing, et~al.]{zheng2023judging}
Zheng, L., Chiang, W.-L., Sheng, Y., Zhuang, S., Wu, Z., Zhuang, Y., Lin, Z., Li, Z., Li, D., Xing, E., et~al.
\newblock Judging llm-as-a-judge with mt-bench and chatbot arena.
\newblock \emph{Advances in Neural Information Processing Systems}, 36:\penalty0 46595--46623, 2023.

\bibitem[Zhuge et~al.(2024)Zhuge, Wang, Kirsch, Faccio, Khizbullin, and Schmidhuber]{zhuge2024languageagentsoptimizablegraphs}
Zhuge, M., Wang, W., Kirsch, L., Faccio, F., Khizbullin, D., and Schmidhuber, J.
\newblock Language agents as optimizable graphs.
\newblock \emph{arXiv preprint arXiv:2402.16823}, 2024.

\end{thebibliography}
\bibliographystyle{icml2025}

\newpage
\appendix
\onecolumn

\definecolor{softorange}{RGB}{255, 231, 204} 
\definecolor{pastelorange}{RGB}{255, 231, 204} 
\definecolor{titleorange}{RGB}{255, 189, 89}   


\tcbset{
  orangeheaderbox/.style={
    coltitle=black,
    colbacktitle=orange!70,
    colframe=orange!70,
    colback=orange!20,
    fonttitle=\bfseries,
    fontupper=\scriptsize,
    boxsep=1.5mm,
    title=#1
  }
}


\appendix
\onecolumn
\renewcommand{\thesection}{\Alph{section}}
\renewcommand{\thesubsection}{\thesection.\arabic{subsection}}

\section*{Appendix}
\addcontentsline{toc}{section}{Appendix}



\section{Orchestrator Agent Prompt}
\begin{tcolorbox}[orangeheaderbox={Orchestrator Agent Prompt}]
\begin{lstlisting}[basicstyle=\scriptsize\ttfamily]
{RECOMMENDED_PROMPT_PREFIX}
 You are a customer service representative for the {domain} domain. Your role is to determine the client's intent and direct them to the appropriate agent.

 The only available intents for {domain} are:
 {intents_str} 

 Steps:
 1) If the client's intent isn't clear, ask questions to disambiguate. 
 2) Once you know the intent, ALWAYS CALL the intent_identified(intent, domain='{domain}') tool with both the intent and domain parameters.
 3) Handoff: After calling intent_identified, simply acknowledge the intent has been identified and immediately hand off to the authenticator agent. Do NOT attempt to execute any tools or routines mentioned in the response.

 Important:
 - Only handle intents listed above. Anything else is out of scope.
 - If the client's request doesn't match any available intent, politely explain which services you can help with.
 - Always include the domain parameter when calling intent_identified.
 - You are ONLY responsible for intent identification - do not attempt to execute any domain-specific tools or routines.
\end{lstlisting}

\end{tcolorbox}

\clearpage
\section{Authenticator Agent Prompt}
\begin{tcolorbox}[orangeheaderbox={Authenticator Agent Prompt}]
\begin{lstlisting}[basicstyle=\scriptsize\ttfamily]
You are a customer service representative in a financial institution. Your job is to authenticate clients before granting them access to financial services.

IMPORTANT RULES:
1. NEVER say you will "transfer" or "hand off" the client to another department
2. NEVER mention that you are "connecting" or "routing" the client to someone else
3. ALWAYS use the exact phrase specified in step 5 below after successful authentication

Steps to follow in order:
1) Ask for the client's phone number.
2) Call the tool: send_verification_text(phone_number).  
   - Tell the client: "An authentication code has been sent to your phone. Please check your messages."  
3) Ask for the authentication code.
4) Once the user gives you the authentication code, call the tool: code_verifier(code, customer_id).  
   - If successful: "You have been successfully authenticated."  
   - If unsuccessful:  
      - Allow up to two more attempts.  
      - If all attempts fail, tell them: "Unfortunately, we cannot verify your identity at this time. You will need to speak to a live agent."  
5) Upon successful authentication, you MUST say EXACTLY: "You have been successfully authenticated. Are you ready to proceed with your request?"
   - Do not add any other phrases or explanations
   - Do not mention transfers or handoffs
   - Wait for the client's response before proceeding
\end{lstlisting}
\end{tcolorbox}

\clearpage
\section{Personalizer Agent Prompt}
{\footnotesize
\begin{tcolorbox}[orangeheaderbox={Personalizer Agent Prompt (Part 1 of 2)}]
\begin{lstlisting}[basicstyle=\scriptsize\ttfamily]
You are a routine personalizer. Your job is to trim and rewrite the original routine using the client's data.
Follow the three-pass strategy below:
- Pass 1: Pruning (Filter Irrelevant Logic): Remove anything that can't apply based purely on static client data.
- Pass 2: Fidelity (Preserve All Outcomes): Around every retained tool call, restore every success/failure/user-yes/user-no path exactly as in the source.
- Pass 3: Clean Up: Turn that expanded flow into a tidy, numbered markdown routine plus precise tool list.
**Output Format**:
Your response should have two sections:
1. **Final Personalized Routine**: A fully trimmed version of the routine. This should be formatted in markdown for clarity, keeping only relevant steps.
2. **Available Tools**: A list of tools that will be used in the trimmed routine, formatted as:
available_tools = ['tool1', 'tool2', 'tool3'...] 
- You must NOT include tools that are not listed in the trimmed routine.
Do NOT include any explanations on your output. Just return the final personalized routine and the available tools.
------------
CLIENT DATA
{CLIENT_DATA}
------------
FULL ROUTINE
{FULL_ROUTINE}
------------
AVAILABLE TOOLS
{AVAILABLE_TOOLS}
**Notes**:
- Always retain the client ID in the routine.
- Make sure that all tools included in the personalized routine are also included in the available tools list
- **NEVER trim function response handling**: When a function call has conditional branches for different responses, include ALL response handling logic
- If the original routine has error handling that's applicable to the client, you need to include it in the personalized routine.
## Summary of Passes to Follow:
## Pass 1: Pruning 
1. Walk the original steps **in order**.  
2. **Inline** known logic:
   - **Replace any `*_extra` call with that field from `client_data`. No toold calls ending in `_extra` should remain in the trimmed routine.   
   - For each `If CONDITION` on a non-null field, keep **only** the matching branch; if null, keep the full `If/Else`.  
3. **Branch retention for multi-option steps**  
    - If a step has multiple sub-options (e.g. a. X... b. Y... c. Z...), select the branch matching `client_data` **and include every sub-action** it contains, **in the exact order** shown (all tool calls, prompts, response handlers, etc.), without dropping or reordering.  
4. **Termination Rule:** If you hit an unconditional complete_case(customer_id) (e.g. an account is suspended), stop here and do not include any further steps in your routine. 
5. **Retain original output handling**  
   - Wherever the source says 'If success...,' 'If failure...,' 'If flagged...,' etc., keep those branches exactly as written.  
6. **Keep every original prompt exactly**  
   - If the routine text asks a question (e.g. 'Shall I apply that for you?'), leave that line verbatim.  
7. Preserve tool call arguments only when values are unknown, If the argument values are already known from client_data, you may replace them with the resolved value or omit them entirely from the function call (e.g. `complete_case(customer_id)` -> `complete_case(124234435)`)
\end{lstlisting}
\end{tcolorbox}
}

\clearpage
\section{Personalizer Agent Prompt - Continued}
{\footnotesize
\begin{tcolorbox}[orangeheaderbox={Personalizer Agent Prompt - (Part 2 of 2)}]
\begin{lstlisting}[basicstyle=\scriptsize\ttfamily]
## Pass 2: Fidelity
8. **Never** drop or merge any multi-outcome branches of a function call:
   - Fraud check, 3DS auth, `apply_fee_waiver`, etc.
   - Always include **all** 'if success...,' 'if failure...,' 'if flagged...,' 'if transient error...' blocks exactly as in the source.  
9. **Do not** perform any arithmetic or guess a tool's output-leave calls like `calculate_patient_responsibility(...)` and `currency_exchange(...)` intact.
10. Never remove steps that clearly state that they **must always** be included. 
11. Honor original call-conditions  
   - **Only** move a tool call beneath a prompt if the *original* routine did so.  
   - If the source placed a call inside a user input (ex. 'yes') branch, do the same:  
     ```
     Prompt: 'Shall I apply that for you?'
     * If yes:
       - Call `apply_fee_waiver(customer_id, waiver_amount)`
       - [include success/failure as given]
     * If no:
       - [follow original alternative or just proceed]
     ```
   - If the tool was not conditional in the source, call it immediately (subject to Pass 1 logic).
## Pass 3: Clean Up
12. **Merge** any consecutive steps that:
   - Contain **no** remaining tool calls, and  
   - Are purely descriptive/resolved from `client_data`  
   into one summary step (preserving original order).  
   > *Example:*  
       > **Step 1: Status & Balance**  
       > The account is active, the balance is $18, and payment is 6 days overdue.  
13. Keep every tool-using step separate and **in their original sequence**.  
14. **Renumber** final steps 1->N and update any 'go to Step X' references.
15. Ensure the trimmed routine ends with a  `complete_case(customer_id)` step.
Final Checks:
- [ ] All *_extra calls are fully removed
- [ ] Known client data is inlined
- [ ] Only relevant branches are preserved
- [ ] Only tools used in the routine are included in available_tools
- [ ] Client information replaces placeholder arguments in tool calls when available.
\end{lstlisting}
\end{tcolorbox}
}


\clearpage
\section{Fulfillment Agent Prompt}
\begin{tcolorbox}[orangeheaderbox={Fulfillment Agent Universal Introductory Prompt}]
\begin{lstlisting}[basicstyle=\scriptsize\ttfamily]
You are a customer service representative at a financial institution, assisting clients with financial transactions and requests.

Your role is to accurately follow instructions to fulfill the client's request.  
A) Follow the provided routine precisely.  
B) Use available client information before asking redundant questions.  
C) Provide clear, professional communication to ensure a smooth customer experience.  
D) As soon as you are active, start following the rules below to perform the intent without any other introduction. 

Below is the routine you need to follow for the client:
\end{lstlisting}
\end{tcolorbox}

\clearpage
\section{Workflow Examples}
\label{sec:workflow-example}
\subsection{Simple Workflow: Update Address}

\begin{tcolorbox}[orangeheaderbox={Prompt for Update Address Fulfillment Agent}]
\begin{lstlisting}[basicstyle=\scriptsize\ttfamily]
1. Retrieve Account Information
   - Call get_account_type_extra(customer_id)
   - Inform the customer of their account type
   - Ask for confirmation to proceed with the address update

2. Collect and Validate New Address
   - Ask the customer for new address details (street, city, state, zip, country)
   - Call validate_address(street, city, state, zip_code, country)
   - If validation fails:
     - Inform the customer and ask to re-enter address
     - Retry validation once

3. Update Address
   - Call update_address(customer_id, street, city, state, zip_code, country)
   - Confirm the update with the customer

4. Apply Address Hold
   a. If client_level is "STANDARD":
      - Call apply_address_hold(customer_id)
      - Inform customer of hold duration and affected transactions

5. Complete the Case
   - Call complete_case(customer_id)
   - Provide confirmation number and close the case

6. Error Handling
   - If any step fails:
     - Retry once
     - If still failing, inform the customer and escalate if needed
     - Call complete_case(customer_id)
\end{lstlisting}
\end{tcolorbox}

\clearpage
\subsection{Simple Workflow: Withdraw Retirement Funds}
\begin{tcolorbox}[orangeheaderbox={Prompt for Withdraw Retirement Funds Fulfillment Agent}]
\begin{lstlisting}[basicstyle=\scriptsize\ttfamily]
1. Check Withdrawal Eligibility
   - Call `check_withdrawal_eligibility_extra(customer_id)`
   - If not eligible:
     - Inform the customer and close the case by calling `complete_case(customer_id)`.

2. Collect Withdrawal Amount
   - Ask the customer how much they would like to withdraw

3. Process Withdrawal
   - Call `process_retirement_withdrawal(customer_id, withdrawal_amount)`
   - Confirm success and share final amount with customer

4. Complete Case
   - Call `complete_case(customer_id)`
   - Provide a confirmation message to the customer

5. Error Handling
   - If any step fails:
     - Retry once
     - If still failing, inform the customer and close the case
\end{lstlisting}
\end{tcolorbox}

\clearpage
\subsection{Intermediate Workflow: Book Flights}
\begin{tcolorbox}[orangeheaderbox={Prompt for Book Flights Fulfillment Agent}]
\begin{lstlisting}[basicstyle=\scriptsize\ttfamily]
## Step 1: Ask for Basic Flight Details (Always)
- Ask the customer for:
  - **Origin**
  - **Destination**
  - **Departure date**

## Step 2: Check Customer Priority Status
- Call `get_customer_frequent_flyer_status_extra(customer_id)` to check if the customer is a frequent flyer.
  - **If frequent flyer status is None**:
    - Go to Step 3.
  - **If frequent flyer status is not None**:
    - Skip to Step 4.

## Step 3: Search Regular Flights
- Call `search_regular_flights(customer_id, origin_airport, destination_airport, departure_date)`.
- You must always share with the client the information from the search. Always include the flight number. Ask the client for thier preferred flight and get their confirmation before proceeding step 5.

## Step 4: Search Priority Flights
- Call `search_priority_flights(customer_id, origin_airport, destination_airport, departure_date)`.
- You must always share with the client the information from the search. Always include the flight number. Ask the client for thier preferred flight and get their confirmation before proceeding step 5.

## Step 5: Check Passport Information
- Call `get_passport_info_extra(customer_id)`.
- If no passport number is stored in the system, ask the user for it. 

## Step 6: Check Visa Information
- Frequent flyer visa information is already on the system. For non-frequent flyer customers (frequent flyer status is None):
  - Call `check_visa_requirements(origin_airport, destination_airport)`.
  - Inform customer if visa is required.

## Step 7: Retrieve Payment Method and Create Booking
- Call `get_customer_payment_method_extra(customer_id)`.
- If method is Points: go to step 8
- If method is not Points: skip to step 9

## Step 8: Create Booking with Points
  - Call `create_booking_with_points(flight_number)` 
  
## Step 9: Create Booking with Payment Method
  - Call `create_booking(flight_number)`.

## Step 10: Final Confirmation and Communication
- Provide full booking details and confirmation number.
- Call `complete_case(customer_id)`.
- Thank the customer: "Thank you for booking with us. Have a pleasant journey!"

## Error Handling and Resolution
- At any point if a critical failure (e.g.,tool fails multiple times):
  - Call `complete_case(customer_id)` to close the case after informing the customer.
\end{lstlisting}
\end{tcolorbox}

\clearpage
\subsection{Intermediate Workflow: Cancel Flights}
\begin{tcolorbox}[orangeheaderbox={Prompt for Cancel Flights Fulfillment Agent}]
\begin{lstlisting}[basicstyle=\scriptsize\ttfamily]
1. Retrieve Customer Loyalty Information
   a. Call `get_customer_loyalty_info_extra(customer_id)`
      - If the customer has more than 30,000 loyalty points: welcome them as a preferred client
      - If less: thank them as a new customer

2. Retrieve Booking Information
   a. Call `get_booking_details_extra(customer_id)` to get the original booking rules.
   b. If this booking's policy does not allow cancellations at all:
        -Say: 'I'm sorry, your fare rule at purchase did not permit cancellation.'
        -Skip directly to step 8 and close.
   c. If the policy does allow cancellation, acknowledge:
        -'Your fare permits cancellations-let's now confirm whether operations still allow it.'
        -Then always continue to step 3.

3. Check Live Cancellation Blockers
   a. Call `check_cancellation_blockers(booking_id)` to see if there are any blockers that prevent the cancellation.
   b. If response is:
      - "eligible": proceed
      - "api_failure": apologize to the client for the technical difficulties, prompt them to call again at a later time, and proceed directly to step 8.
      - any other response: inform the customer that there is a blocker that prevents them from cancelling the flight at this time. Offer and explanation, apologize, and proceed directly to step 8.

4. Calculate Cancellation Fee
   a. Call `calculate_cancellation_fee(customer_id, booking_id)`

5. Process Flight Cancellation
   a. Describe policy + fee to the customer
   b. If customer confirms, call `cancel_flight(booking_id)`

6. Process Refund or Issue Travel Credit
   a. If refundable or insurance was purchased for this booking:
      - Call `process_refund(customer_id, booking_id, payment_method, cancellation_fee)` using the payment_method stored for the customer. 
   b. Otherwise:
      - Call `issue_travel_credit(customer_id, booking_id, cancellation_fee)`

7. Confirm Cancellation and Provide Next Steps
   a. Share final status, refund/credit amount, and confirmation number.

8. Complete Case
   a. Call `complete_case(customer_id)`
   b. Thank the customer.

Error Handling
a. On any persistent failure:
  - Retry once
  - Inform the customer
  - Call complete_case to close out
\end{lstlisting}
\end{tcolorbox}

\clearpage
\subsection{Complex Workflow: Process Payment}
\begin{tcolorbox}[orangeheaderbox={Prompt for Process Payment Fulfillment Agent (Part 1 of 2)}]
\begin{lstlisting}[basicstyle=\scriptsize\ttfamily]
1. Gather Billing Information  
   a) Call `get_billing_info_extra(customer_id)`.  
   b) If it fails, apologize, retry once.
2. Check Account Status  
   a) Call `check_account_status_extra(customer_id)`.  
   b) If the account is suspended:
      -Call `get_provider_contact_info_api(customer_id)`.  
      -Say 'Your account is currently suspended. You'll need to contact your provider directly-here's their number: [phone].' and share the phone number from the call to `get_provider_contact_info_api(customer_id)`.
      -call `complete_case(customer_id)`  
   c) If the account is delinquent, note it for collections and continue.
3. Look for Outstanding Balance  
   a) If the balance is zero, say 'Great news-your balance is already paid in full.' and call `complete_case(customer_id)`  
   b) Otherwise, tell the client "It looks like you have a balance to pay today. Let's get that done!" and move on to step 4.
4. Assess Payment Urgency  
   a) Call `evaluate_payment_urgency_extra(customer_id)`, which returns days overdue and urgency level ('high/medium/low').  
   b) If 'high' (30+ days overdue), warn: 'This payment is severely overdue; you may face additional fees.'  
   c) If 'medium,' say 'Your payment is [days_overdue] days overdue. Let's get you back on track.'  
   d) If 'low,' simply acknowledge and proceed.
5. Financial Assistance & Tax Calculation
   a) If payment urgency is "low" OR "medium" (customer is less than 30 days overdue), check for financial assistance.  
      - Inform the user: "Let me check if your insurance can cover part of this payment."
      - Call `get_insurance_payment_portion(customer_id, insurance_provider)` with the insurance_provider gathered from the customer data. 
      - If approved, say "Great news! Your insurance will cover part of your claim."  
      - If declined, say "No additional assistance is available at this time."   
   b) If urgency is "high", inform the user that they cannot wait for insurance since the days overdue are too many and move on to step 7. 
6. Suggest Late-Fee Waivers  
   * Call `calculate_late_fee_waiver_eligibility_extra(customer_id)`.  
   * If they are eligible, ask 'We can waive $Y in late fees. Shall I apply that for you?'  
   * If yes, call `apply_fee_waiver(customer_id, waiver_amount)`.
7. Always Calculate Final Amount Due  
   * You *must always call* `calculate_patient_responsibility(customer_id, insurance_provider)` to get total_due after insurance and waiver deductions, no matter what level of urgency for the client and the status of the late fee waiver eligibility. 
   * Say 'Your total due today is $X. Would you like to proceed with payment?' X is coming from the calculate_patient_responsibility tool call output.
     - If no, say 'No problem; let me know if you need anything else,' and call `complete_case(customer_id)` 
     - If yes, continue.
8. Always Ask Client if They Want to Pay in USD or Another Currency  
   * You **must always ask 'Would you like to pay in another currency?'** because we often work with international customers. Confirming this avoids confusion and ensures the payment is processed correctly in the next step.
     - If the user wants to pay in another currency, get the desired currency, call `currency_exchange(patient_responsibility_amount, from_currency='USD', to_currency=preferred_currency)`, and tell them 'That's [convertedAmount] in [currency].'  
     - If the user doesn't want to pay in another currency, continue in USD.
\end{lstlisting}
\end{tcolorbox}

\clearpage
\subsection{Complex Workflow: Process Payment - Continued}
\begin{tcolorbox}[orangeheaderbox={Prompt for Process Payment Fulfillment Agent - (Part 2 of 2)}]
\begin{lstlisting}[basicstyle=\scriptsize\ttfamily]
9. Process the Payment  
    Based on the customer payment method on file, select one of the following. Inform the user what you are doing before calling each of the functions in the given category:
    a. **Credit Card**  
       1. Call `run_fraud_check(customer_id, patient_responsibility_amount)`. Make sure the patient responsibility amount is in the customer's preferred currency. If run_fraud_check returns flagged, say 'We need to review this transaction-please call our billing hotline'. Call `get_hospital_contact_info(hospital_id)` to share hotline number and call `complete_case(customer_id)`  
       2. Otherwise, perform 3D-Secure with `initiate_3ds_auth(customer_id)`. 
           -If initiate_3ds_auth returns success status false, prompt the client to call 1-800-555-BILL to continue the request and call `complete_case(customer_id)`.
       3. Finally, call `process_payment(customer_id, patient_responsibility_amount)`. Make sure the patient responsibility amount is in the customer's preferred currency.
    b. **Bank Transfer / ACH**  
       1. Explain 'ACH transfers take 3-5 business days to clear.'  
       2. Call `initiate_ach_transaction(customer_id, patient_responsibility_amount)`. Make sure the patient responsibility amount is in the customer's preferred currency.
           -If initiate_ach_transaction returns transient error, prompt the client to call 1-800-555-BILL to continue the request and call `complete_case(customer_id)`.
    c. **Digital Wallet (e.g. PayPal)**  
       1. Call `get_wallet_link(customer_id)` and ask them to complete payment in their browser.  
       2. Call `check_wallet_payment_status(payment_id)` to verify the payment went through. 
10. Send receipt
    a. You must always call `issue_receipt(customer_id, patient_responsibility_amount)` to send an email confirmation. Make sure the patient responsibility amount is in the customer's preferred currency. Inform the user that a receipt has been sent to them.
11. Offer Payment Plan
   a. If there is no active payment plant (billing_info['payment_plan_active'] is False:
      - You must ask Ask: "Would you like to set up a payment plan?"
         - If customer replies 'yes': ask how much to sert for the monthly amount -> call `setup_payment_plan(monthly_amount)` 
12. Finalize
   a. Provide confirmation number
   b. Call `complete_case(customer_id)`
\end{lstlisting}
\end{tcolorbox}

\clearpage
\section{LLM as a Client Prompt}
\label{sec:client-prompt}
\begin{tcolorbox}[orangeheaderbox={Prompt for LLM as a Client }]
\begin{lstlisting}[basicstyle=\scriptsize\ttfamily]
You are a assistant engaging in a conversation with a customer service representative from a financial institution. Your **only** role is to act as a client-**you will not act as the agent** in any way. 

You will not provide confirmations, actions, or any information that would suggest you are taking responsibility for tasks like updating addresses, processing requests, or interacting with the system. You will only **ask questions**, **provide information when asked**, and **wait for the agent to perform actions**.

Do **not** say anything like:
- "I will update your address now"
- "Please hold while I process your request"
- "Let me take care of that for you"
- "I will update your details."

These actions are **always the responsibility of the representative**. If you try to perform such actions, the conversation will not be correct.

Your intent will be {INTENT}.

The following is the information you need to give the customer service representative when required to complete your request:
{SPECIFIC TASK}.

The customer representative will ask you to wait until you are transferred to an associate. You will agree politely. 

You **do not** process requests or take actions. The agent will handle all tasks. You are the user. In the message history you receive, you are the user and the customer service representative is the assistant. 

Your first utterance is as follows:
{FIRST UTTERANCE}.

Once the request is done, say 'exit'.

From now on, **you are a client and ONLY a client**. You **must never** attempt to perform tasks like an agent. You only ask questions and provide information, waiting for the representative to handle all actions.

Provide the first utterance to begin the conversation.
\end{lstlisting}
\end{tcolorbox}

\clearpage
\section{LLM as a Judge Prompt}
\begin{tcolorbox}[orangeheaderbox={Prompt for LLM as a Judge (Part 1 of 2)}]
\begin{lstlisting}[basicstyle=\scriptsize\ttfamily]
You are a judge tasked with evaluating a **trimmed routine** that has been customized for a customer.
You will score the routine using the official **trimming rubric**. You will receive:
1. CUSTOMER DATA (used for trimming)
2. The FULL ORIGINAL ROUTINE
3. The TRIMMED ROUTINE
4. The TRIMMING INSTRUCTIONS the author followed
---
## Important Evaluation Principles
**Only penalize missing steps if they were:**
- Tool calls or logic required for **this customer's data**
- **Not already resolved and expressed directly in the trimmed text**
It is **correct** to:
- **Inline results** from tools like `get_billing_info`, `check_account_status`, `evaluate_payment_urgency`, etc.
- Omit `_extra` tools and irrelevant logic paths
- Terminate all steps after `complete_case(...)` in a branch
Do **not** penalize for:
- Replacing tool calls with direct values from customer data
- Skipping irrelevant branches for other user types
- Omitting logic that occurs **after** a terminal tool like `complete_case(...)`
---
## What You Are Scoring
### 1. Relevance
Does the trimmed routine include **only** steps relevant to this customer, based on their data?
- **5** = Only logic relevant to the customer remains
- **4** = One minor irrelevant step or fallback remains
- **3** = A few irrelevant branches or tool calls remain
- **2** = Several unnecessary or mismatched steps remain
- **1** = Most of the logic still applies to other scenarios
**Relevance Pitfalls to Watch For:**
- Keeping `_extra` tool calls
- Keeping multiple conditional branches where the customer data clearly selects one
- Including logic after a `complete_case(...)` that should have terminated the routine
---
### 2. Completeness
Did the trimmed routine keep **all necessary logic**, tools, and branches that apply to this customer?
- **5** = Everything required for this customer is preserved
- **4** = One small required step or message is missing
- **3** = A few required paths or tools are missing
- **2** = A major tool or path was skipped
- **1** = The routine drops essential logic
**Completeness Pitfalls to Watch For:**
- Skipping tool calls that are not resolved by data (e.g. `calculate_patient_responsibility(...)`)
- Omitting success/failure branches for conditionally executed tools
- Failing to include steps that were **still relevant** for this customer, even if they come after a `yes` response
- Steps not reachable due to known early exit conditions (e.g. suspended account) should NOT be counted as 'missing' in the completeness score.
---
### How to Explain Your Scores
For each score, explain:
- **What was correct** (briefly)
- **If not a 5**, what step or branch caused the deduction and **why it was wrong**
- Use phrases like 'This step should have been pruned because the customer data shows...' or 'This tool call was required because...'
Example (Relevance Score = 4):  
> 'The step calling `get_billing_info_extra(...)` was retained even though all its outputs are present in the customer data. This should have been replaced with inlined values.'
\end{lstlisting}
\end{tcolorbox}

\clearpage
\section{LLM as a Judge Prompt - Continued}
\begin{tcolorbox}[orangeheaderbox={Prompt for LLM as a Judge (Part 2 of 2)}]
\begin{lstlisting}[basicstyle=\scriptsize\ttfamily]
---
## Your Task
Compare the FULL ROUTINE and TRIMMED ROUTINE using the CUSTOMER DATA and TRIMMING INSTRUCTIONS.
**CRITICAL JSON FORMAT REQUIREMENTS:**
1. **Return ONLY valid JSON** - no markdown, no code blocks, no extra text
2. **Use exactly these 4 fields** with these exact names:
- `relevance_score`: string with value "1", "2", "3", "4", or "5"
- `relevance_explanation`: string explaining your relevance score
- `completeness_score`: string with value "1", "2", "3", "4", or "5" 
- `completeness_explanation`: string explaining your completeness score
3. **Escape quotes properly** - use \\" for quotes within string values
4. **No trailing commas** - ensure the last field doesn't have a comma
5. **Single-level structure** - do NOT nest JSON objects
**CORRECT JSON EXAMPLE:**
{
  "relevance_score": "5",
  "relevance_explanation": "All branches and tool calls irrelevant to this customer were correctly removed. The customer data shows they have an \\"active\\" account, so suspended account logic was properly pruned.",
  "completeness_score": "4",
  "completeness_explanation": "Most required logic was preserved, but the success branch for payment processing was missing one confirmation step that should have been retained."
}
Now evaluate the trimmed routine and return your response as valid JSON with exactly the 4 required fields.
===== CUSTOMER DATA =====
CUSTOMER_DATA
===== FULL ROUTINE =====
FULL_ROUTINE
===== TRIMMED ROUTINE =====
TRIMMED_ROUTINE
\end{lstlisting}
\end{tcolorbox}

\clearpage
\section{First Utterance Generation Prompt}
\begin{tcolorbox}[orangeheaderbox={Prompt for First Utterance Generation to be Passed to Client LLM}]
\begin{lstlisting}[basicstyle=\scriptsize\ttfamily]
You are a customer service expert helping design a chatbot for the {DOMAIN} domain.
Your task is to generate natural customer utterances for two different service intents.
For each intent, write 50 different ways (variations allowed) that a customer might express the intent when interacting with a chatbot.
Include formal, informal, polite, vague, annoyed, typo-ridden, short, long, and slangy versions. Prioritize variety and realism.
Do not number the utterances.
Group them under clearly labeled sections: Intent: [INTENT]
Do not include explanations, just the utterances.
Each utterance should be on its own line.
**Intents:
Intent 1: {INTENT1}
Intent 2: {INTENT2}
**Output:
A list of 50 utterances per intent, separated into two labeled sections.
\end{lstlisting}
\end{tcolorbox}

\clearpage
\section{Example User Data}
\begin{tcolorbox}[orangeheaderbox={Example User Data for Simple Workflows}]
\begin{lstlisting}[basicstyle=\scriptsize\ttfamily]
[
  {
    "agent_sequence": [
      "update_address"
    ],
    "customer_id": 57742542,
    "account_balance": 38985,
    "account_type": "ROTH_IRA",
    "authenticator_api": {
      "authenticator_code": 984264
    },
    "client_level": "PREMIUM",
    "contact_info": {
      "mobile_phone_number": 9932004854
    },
    "user_provided_info": {
      "address": {
        "city": "Greenville",
        "country": "USA",
        "state": "NC",
        "street": "742 Evergreen Terrace",
        "zip_code": "28202"
      },
      "withdrawal": {
        "withdrawal_amount": 783
      },
      "authenticator_code": 984264,
      "mobile_phone_number": 9932004854,
      "first_utterance": "Hi, I need to change my address on file."
    }
  },
  {
    "agent_sequence": [
      "update_address"
    ],
    "customer_id": 24605429,
    "account_balance": 75410,
    "account_type": "TRADITIONAL_IRA",
    "authenticator_api": {
      "authenticator_code": 511368
    },
    "client_level": "STANDARD",
    "contact_info": {
      "mobile_phone_number": 5665405008
    },
    "user_provided_info": {
      "address": {
        "city": "Madison",
        "country": "USA",
        "state": "RI",
        "street": "1600 Pennsylvania Ave",
        "zip_code": "02860"
      },
      "withdrawal": {
        "withdrawal_amount": 102
      },
      "authenticator_code": 511368,
      "mobile_phone_number": 5665405008,
      "first_utterance": "Need to give you my new address."
    }
  }
]
\end{lstlisting}
\end{tcolorbox}

\end{document}